%% file: mainv4.tex
\documentclass[journal]{IEEEtran}
\input{packages}
\begin{document}
\title{DIAL: Deep Interactive and Active Learning for Semantic Segmentation in Remote Sensing}

\author{Gaston~Lenczner, 
        Adrien~Chan-Hon-Tong, %~\IEEEmembership{Member,~IEEE,}
        Bertrand~Le~Saux, %~\IEEEmembership{Member,~IEEE,}
        Nicola~Luminari, 
        and~Guy~Le~Besnerais% <-this % stops a space
\thanks{Gaston Lenczner, Adrien Chan-Hon-Tong, and Guy~Le~Besnerais are with ONERA/DTIS, Universit{é} Paris-Saclay, FR-91123 Palaiseau, France}% <-this % stops a space
\thanks{Gaston~Lenczner and Nicola~Luminari are with Alteia, FR-31400 Toulouse, France (email: gaston.lenczner@alteia.com)}
\thanks{Bertrand Le Saux is with ESA / ESRIN $\Phi$-lab, IT-00044 Frascati (Roma), Italy (email: bertrand.le.saux@esa.int)}
}

\markboth{Journal of \LaTeX\ Class Files,~Vol.~X, No.~X, December~2021}%
{Shell \MakeLowercase{\textit{et al.}}: Bare Demo of IEEEtran.cls for Journals}

\maketitle

\begin{abstract}
We propose in this article to build up a collaboration between a deep neural network and a human in the loop to swiftly obtain accurate segmentation maps of remote sensing images. In a nutshell, the agent iteratively interacts with the network to correct its initially flawed predictions. Concretely, these interactions are annotations representing the semantic labels. Our methodological contribution is twofold. First, we propose two interactive learning schemes to integrate user inputs into deep neural networks. The first one concatenates the annotations with the other network's inputs. The second one uses the annotations as a sparse ground-truth to retrain the network. Second, we propose an active learning strategy to guide the user towards the most relevant areas to annotate. To this purpose, we compare different state-of-the-art acquisition functions to evaluate the neural network uncertainty such as ConfidNet, entropy or ODIN. Through experiments on three remote sensing datasets, we show the effectiveness of the proposed methods. Notably, we show that active learning based on uncertainty estimation enables to quickly lead the user towards mistakes and that it is thus relevant to guide the user interventions.  
\end{abstract}


\IEEEpeerreviewmaketitle

\input{sectionsv3/intro}
\input{sectionsv3/methodology}
\input{sectionsv3/experiments}
\input{sectionsv3/conclu}

\appendices

\ifCLASSOPTIONcaptionsoff
  \newpage
\fi

\bibliographystyle{IEEEtran}
%\bibliography{bibli}

\input{sectionsv3/supp_material}
\end{document}

%% file: packages.tex
\usepackage{graphicx}
\usepackage{epsfig}
\usepackage{amsfonts}

\usepackage{amsmath}
\usepackage{hyperref}
\hypersetup{colorlinks=true, allcolors=blue, urlcolor=purple}
\usepackage{soulutf8}%
\usepackage[colorinlistoftodos]{todonotes}
\usepackage{color}
\usepackage{xargs}%
\usepackage{float}
\usepackage{wrapfig}
\usepackage{multirow}
\usepackage[symbol]{footmisc}
\usepackage{subcaption}
\usepackage{booktabs}
\usepackage{array}

\DeclareMathOperator*{\argmax}{\arg\!\max}
\DeclareMathOperator*{\argmin}{\arg\!\min}
\usepackage{mathtools, stmaryrd}
\usepackage{xparse} \DeclarePairedDelimiterX{\Iintv}[1]{\llbracket}{\rrbracket}{\iintvargs{#1}}
\NewDocumentCommand{\iintvargs}{>{\SplitArgument{1}{,}}m}
{\iintvargsaux#1} %
\NewDocumentCommand{\iintvargsaux}{mm} {#1\mkern1.5mu..\mkern1.5mu#2}
\usepackage{stmaryrd}

\newcolumntype{?}{!{\vrule width 1pt}}
\usepackage{colortbl}
\definecolor{Gray}{gray}{0.85}
\newcolumntype{a}{>{\columncolor{Gray}}c}

%% file: sectionsv3/intro.tex
\section{Introduction}
\label{intro}
\subsection{Context}

\color{brown}

\color{black}

% \IEEEPARstart{S}{emantic}
\IEEEPARstart{S}{emantic} segmentation, the task of classifying an image at the pixel level, is extremely important in remote sensing and is addressed with deep neural networks for a variety of applications such as land-cover mapping~\cite{costa2018supervised}, change detection~\cite{daudt2018fully} or farmland monitoring~\cite{chiu2020agriculture}. This task is intrinsically complex and, while deep neural networks can be very effective, they are still prone to failure. Indeed, even on academic benchmarks~\cite{maggiori2017dataset,Cordts2016Cityscapes}, current state-of-the-art methods often require specific architectures and fine-tuning to obtain high performances but still imperfect results. Moreover, it often gets even more tedious on "real-life" datasets due to different factors such as domain adaptation between train and test data inherent to remote sensing data (different weather, geographical areas, types of sensors, cloud shadows, etc) or the difficulty to have access to large training annotated datasets for every specific business application, even though lots of efforts are made by the community in this direction~\cite{castillo2020semi,chen2019}. Hence, the uncertainty about the quality of the results of neural networks often makes their deployment complicated. Human intervention may then be necessary. Precisely, we are thinking of two scenarios representative of real situations. First, the \textit{Refinement} use-case, when the user aims to fix errors within a single dataset and thus to improve the performances of the model on the test data. Second, the \textit{Domain Adaptation} use-case, when the user wants to fix on a new dataset the errors of a model pre-trained on a previous dataset. 

A possible way to address these problems comes with Interactive Learning (IL)~\cite{schroder2000interactive,le_saux_iros2013}. This consists in adding a human in the loop to work in synergy with a learning algorithm to train it, fine-tune it or adapt it to user inputs. Compared to classically supervised algorithms, IL algorithms must also interface smoothly with the human user. This constraint is particularly challenging with deep neural networks due to their typical high number of parameters and long training time.  One step further than IL, Active Learning~\cite{settles2009active} (AL) searches in pools of unlabeled data for examples which are the more able to lead the model to a better classification. These examples, defined as \textit{queries}, are then labeled by the user and incorporated in the training. This thus aims to find the optimal training dataset for the algorithm. To this purpose, active learning methods define \textit{acquisition functions} to estimate either the model uncertainty associated to the samples~\cite{gal2017deep} or their representativeness of the dataset~\cite{sener2017active}. %have the additional constraint to smoothly interface
\input{codes/figures/full}

In this article, we explore IL and AL for semantic segmentation. Indeed, as presented in Figure~\ref{fig:full}, we propose the Deep Interactive and Active Learning (DIAL) framework to interactively refine semantic segmentation maps initially output by a pre-trained neural network. First, it relies on two complementary IL schemes to integrate information provided by a user in deep learning algorithms for semantic segmentation. In a nutshell, the first module uses these annotations to modify at test time the inputs of the network pre-trained to process them while the second one uses them for retraining to modify the weights of the network.
Second, we integrate active learning within our framework and propose to guide the user interventions towards relevant areas to annotate. This additional guidance relies on different uncertainty measures that we compare with respect to our framework. These measures can be simple-yet-effective such as entropy~\cite{shannon1948mathematical} or come from the current state-of-the-art literature such as ODIN~\cite{liang2017principled} or ConfidNet~\cite{corbiere2019addressing}. We extensively evaluate our framework both in the Refinement and in the Domain Adaptation use-cases to well apprehend its potential. We notably show that the first IL module is more suited to correct spatially small mistakes while the second one is more suited for larger ones and that the active process improves the performance compared to the unguided one.

In summary, our major contributions are as follows:
\begin{enumerate}
    \item We propose a general framework for interactive multi-class semantic segmentation in remote sensing.
    \item We show that active learning for area selection allows to speed-up the improvement of the segmentation and reduces the number of required interactions for a given quality. 
    \item We compare different state-of-the-art methods to estimate the algorithm uncertainty within the DIAL framework sketched in Figure~\ref{fig:full}, and show that the model confidence evaluated by ConfidNet and surprisingly the entropy are the most effective, the latter being also the faster one.
    \item We show these techniques consistently improve the quality of a segmentation map, with the greater gain for the domain adaption use-case where it allows to compensate for the lack of training data in the target domain.
\end{enumerate}
This article extends previous work presented in an international conference~\cite{lenczner2020interactive}. It deepens the interactive learning experiments and combines the interactive and active modules within a single framework.

\subsection{Scenario}
We assume the following context for the rest of the paper. A user needs to quickly and accurately semantically segment Earth observation images for one of the two aforementioned use-cases: the Refinement one and the Domain Adaptation one. This user has also access to another annotated database which, depending on the use-case, may or may not belong to the same domain as the targeted images. For the sake of simplicity, the annotated label space must be the same as the targeted one. 

We propose to first train a neural network on the annotated database. Then, the user can use this neural network to make predictions on the target images. If the segmentation result is not accurate enough for the user's requirements, they can interact with the network to refine its predictions. These user interactions come in the form of clicked points on the mislabeled areas and represent their corresponding labels, as chosen by the user. Finally, we propose to also guide the user to the most relevant areas of the images using uncertainty estimations relying on different statistical measures. % the labels of the clicked pixels

We have developed a QGIS\footnote{\href{http://qgis.osgeo.org}{http://qgis.osgeo.org}} plugin available with the code to allow potential users to experience the proposed framework. However, to conduct a large scale evaluation, we have also simulated the user behavior to automatically generate interactions.
Hence, in the rest of this article, we refer both to the synthetic operator and to the potential human user as \textit{the agent}. 

When the agent is simulated to automatically generate the annotations, it samples them in the mistake areas using a comparison between the ground-truth map and the prediction map. It thus necessarily requires a partial access to the ground-truth maps.

To summarize, our framework combines the three following criteria:
\begin{itemize}
    \item \textbf{Semantic segmentation:} The neural network is able to provide accurate semantic segmentation maps without guidance.
    \item \textbf{Interactive Learning:} The neural network can also refine these segmentation maps using the annotations to efficiently fix its mistakes.
    \item \textbf{Active Learning: } It estimates the neural network uncertainty to guide the user towards queries. 
\end{itemize}

\subsection{Related Work}

 \subsubsection{Interactivity in remote sensing} Interactive interpretation of remote sensing data has a long history, partially due to the lack of reference data for training in that field.  Interactivity has been processed by various techniques to enhance data mining tools with relevance feedback capability~: Bayesian modeling of sample distributions was at the core of VisiMine~\cite{aksoy-interactive-KDD04}, and Support Vector Machines (SVMs) were used in~\cite{ferecatu-interactive-TGRS2007}. More recently, boosting and random forests~\cite{breiman2001random} have been the method of choice due to the possibility to train quickly in an incremental manner~\cite{le_saux_iros2013,dos-santos-gosselin-interactive-JSTARS2013,baetens2019validation}. Precisely, ALCD~\cite{baetens2019validation} trains a random forest on user annotations.  With respect to these works, our approach applies deep learning for interactive remote sensing, which is challenging due to the long training time inherent to deep neural networks. Finally, active learning has recently been applied to deep learning in remote sensing~\cite{ruuvzivcka2020deep,kellenberger2019half} to interactively update the models. \cite{kellenberger2019half} addresses detection of extremely rare objects  (e.g. animal detection in aerial images) while \cite{ruuvzivcka2020deep} deals with rare and varied change detection. On our side, we apply active learning to deep learning in the context of segmentation maps refinement. %DIAL can be considered an alternate approach to those ones in the context of segmentation maps refinement.  % Active learning, or in other words looking for examples which are the more able to lead to a better classification, has also been  used for this purpose~\cite{tuia2009,bruzzone-persello-2009} % based on its previous mistakes and on its confidence  to classify the target images

\subsubsection{Interactive Segmentation} Interactive Segmentation intends to interactively segment an image into foreground and background pixels with user annotations. It was initially addressed using graph-cut based methods~\cite{rother2004grabcut} and now mostly by deep neural networks which take as inputs a concatenation of the RGB image and user annotations~\cite{xu2016deep}. In~\cite{kontogianni2020continuous}, the authors use the annotations as sparse ground truth maps to interactively adapt the neural network to a specific object. Multi-class interactive segmentation broadens interactive segmentation to correct multi-class segmentation maps. ~\cite{agustsson2019interactive} proposed a neural network which takes as input a concatenation of the image and the extreme points of each instance in the scene and then lets a user correct the proposed multi-class segmentation using scribbles. We do not assume such extreme point map availability as it is extremely costly to acquire in a remote sensing image with potentially many objects.

\subsubsection{Weakly supervised segmentation} When labels are scarce, training usually boils down to learning the most out of the available labels while leveraging unlabeled data to learn a better inner representation as support. To address weakly supervised semantic segmentation in remote sensing,~\cite{li2021effectiveness} uses image-level labels while~\cite{wang2019weakly} focuses on domain adaptation using bounding boxes in the target space. Close to semantic segmentation,~\cite{caye2019guided} addresses change detection in a weakly supervised setting. Semantic segmentation with point supervision was first proposed by  What's the Point~\cite{bearman2016s} (WTP)  which trains a model from scratch using cross entropy loss on the point labels. Recently and closely related to our work,~\cite{hua2021semantic} proposed FESTA for weak semantic segmentation in remote sensing. It mainly consists in a regularization to train a neural network from-scratch using notably point labels. In our paper, we start from a neural network pre-trained with full supervision instead of starting from scratch. In our retraining component, we also design a regularization suited to our use-cases.

 \subsubsection{Active Learning} Active Learning aims at optimizing the training process of a learning algorithm through an iterative collaboration with a human oracle~\cite{settles2009active}. It makes the algorithm choose from a pool of unlabeled data which ones would be the most relevant to improve itself. Then, the oracle provides the asked labels and the algorithm can learn from it. As it defines how to select the data samples to annotate, the acquisition function is the key differentiating component of these methods. These acquisition functions usually rely on an uncertainty, a representativeness or a diversity score computed directly with the model to select the most relevant samples. Uncertainty methods can rely on simple criteria like entropy~\cite{shannon1948mathematical} or disagreement between ensemble models~\cite{hansen1990neural} to estimate the model's prediction confidence. As uncertainty-based methods do not aim to be representative of the dataset, they can select very similar examples. To address this issue, representativeness-based methods aim to select the samples in order to form a subset as representative as possible of the entire dataset. Addressing this as a core-set approach,~\cite{sener2017active} solves it like the K-center problem using the L2 distance between the activations of the final fully-connected layer of the CNN. Finally, often relying on clustering~\cite{nguyen2004active,demir2010batch}, diversity-based active learning aims to select samples that are as diverse (i.e. different) as possible to reduce the redundancy among the selected samples. In the past decade, active learning has been deeply explored in remote sensing to train algorithms for animal detection~\cite{laroze2018active,kellenberger2019half}, image classification~\cite{tuia2009,bruzzone-persello-2009}, image segmentation~\cite{guo2016superpixel} and recently for change detection~\cite{ruuvzivcka2020deep}. We borrow from active learning techniques to smoothly help the agent to guide our interactive neural network. We focus on uncertainty measures which fit our use-case better than representativeness ones since we aim to easily spot wrong predictions and not to increase our training set. % Finally, ALCD~\cite{baetens2019validation} applies a so-called "supervised active learning" strategy for cloud detection in satellite data which lets the user select the new training samples to annotate.

\subsubsection{Uncertainty in deep neural networks}
Uncertainty quantification, or confidence estimation, is a long-standing problem in machine learning and has many applications such as out-of-distribution~(OoD) samples detection~\cite{liang2017principled}, the decision to trust the model or to defer to a human expertise in fields like healthcare or the detection of new classes in class-incremental learning~\cite{rebuffi2017icarl}. Notably, it can also be used in active learning to determine which samples should be sent to the oracle for annotation. Many methods to estimate the uncertainty in deep neural networks have been recently proposed, and they often fall into one of these four categories. % or even in Bayesian optimization to decide which configuration should be explored % and we draw inspiration from this literature

\paragraph{Softmax probabilities} The first category of methods uses the probabilities from the softmax output space of the neural networks. Indeed,~\cite{hendrycks2016baseline} propose a simple yet strong baseline using the maximum class probability as an uncertainty estimation and apply to outliers detection. However, it is now well-established that softmax probabilities are prone to different issues such as poor calibration~\cite{guo2017calibration} and not fit to differentiate in- from out-of-distribution samples~\cite{hendrycks2016baseline}. To overcome these issues,~\cite{liang2017principled} propose ODIN to detect outliers with a tempered softmax and with adversarial inputs to better distinguish inliers from outliers. Similarly,~\cite{lee2018simple} perturbs their inputs but instead uses the representation space before the softmax layer and the Mahanabolis distance to do the split. 

\paragraph{Model ensembling} Due to its intuitive concept and ease of implementation, another popular class of methods estimate the confidence associated to a sample by measuring the disagreement of different models. This model ensembling can either be explicit and use different models~\cite{beluch2018power} or implicit to be less computationally greedy with one stochastic model using dropout~\cite{gal2016dropout} (MC Dropout) or batch normalization~\cite{ruuvzivcka2020deep}. However, all these methods inherently require several forward propagation and are thus relatively slow, making them not engaging for interactive interpretation. 

\paragraph{Auxiliary models} Other recent approaches design an auxiliary model to learn the uncertainty of the downstream model. While ~\cite{devries2018learning} mostly focuses on OoD detection,~\cite{corbiere2019addressing} addresses failure prediction and proposes ConfidNet, a neural network to predict if the prediction from the downstream network is accurate or not. These methods do not require to retrain the downstream network and can thus be easily plugged into any pre-trained architecture. However, they are computationally heavy and require a new training phase for each new task and model. In remote sensing,~\cite{garcia2020uncertainty} successfully applies the ConfidNet method for land cover segmentation.

\paragraph{Customized loss} Finally, some works design a specific loss to learn the uncertainty directly during training. For instance,~\cite{yoo2019learning} trains a model to predict the loss associated to a prediction and~\cite{moon2020confidence} proposes a loss which regularizes the class probabilities to better estimate uncertainty. These methods are computationally efficient and model agnostic but require a full training from scratch and can't be plugged in a pre-trained model.

We compare different methods from these categories in the present work to optimally guide the agent toward relevant areas to annotate. We focus on the three first categories since an adapted loss is less tailored for our use-case as it would require training new models from scratch.%\bertrand{}{Comment: Here, very good!}

%% file: codes/figures/full.tex
\begin{figure*}[ht]
\centering
\begin{minipage}[t]{.9\linewidth}
\epsfig{figure=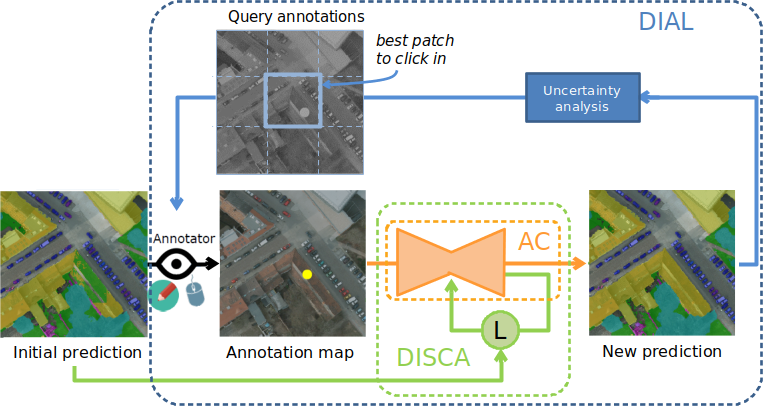,width=1\linewidth}
\end{minipage}
	\caption{Visual representation of DIAL encompassing Deep  Image  Segmentation  with  Continual  Adaptation (DISCA) and Active Learning. The framework starts with an initial prediction that the annotator can annotate with new points (e.g. to fix errors). Three algorithmic mechanisms cooperate to improve the segmentation map: Annotations as Channels (AC) processes jointly image and annotations without retraining, DISCA additionally retrains the model for better adaptation, and DIAL also proposes most informative patches to speed up the interactions. Best viewed in color.} %  Blue corresponds to the initial common process, orange to the DISIR process and green to the DISCA process.% The RGB image and the annotation maps are concatenated before the inferences but not to retrain the network with DISCA.
\label{fig:full}
\end{figure*}

%% file: sectionsv3/methodology.tex
\section{DIAL: Deep Interactive and Active Learning}  

% two complementary algorithms, namely DISIR and DISCA, that we have designed to use agent annotations to interactively guide deep neural networks to refine segmentation maps.
% \gaston{DIAL encompasses three different modules to interactively guide deep neural networks to refine segmentation maps. The first one, DISIR, modifies the neural network input channels to process encoded agent annotations. The second one is based on continual learning to modify the neural network using the same annotations. DISCA is a combination of these two first modules. Finally, since remote sensing images can be extremely large,  DIAL also incorporates an active learning strategy to swiftly guide the agent towards queries representing the most meaningful areas of the image to annotate. We now delve in the details of these three components}{
DIAL encompasses different interactive learning modules and an active learning module to interactively guide deep neural networks to refine segmentation maps and we now delve in the details of these components, which are also illustrated in Figure~\ref{fig:full}.

% \subsection{DISIR: Deep Image Segmentation with Interactive Refinements}
\subsection{Interactive learning components}

The three following interactive learning mechanisms form the Deep Image Segmentation with Continual Adaptation (DISCA) module:

\paragraph{Annotations as Channels (AC)} The neural network takes as input a concatenation of the RGB image and agent annotations, extending ideas of DIOS~\cite{xu2016deep} %,liew2017regional
to multi-class segmentation. At test time, these annotations are initially provided by the agent in the form of clicked points, then encoded using distance transforms into a $N$-dimensional  tensor  where $N$ is  the  cardinal  of  the  label  space. During  the  initial  supervised  training  phase, since the neural network needs to learn how to use clicked points as guidance to enhance its initial predictions, we simply provide points randomly sampled from the ground truth to the network. Image-only inputs are also sampled to train segmentation in a standard way and ensure that the network proposes accurate initial segmentation maps. %they are simply sampled at random from  the  ground-truth

% \subsection{DISCA: Deep Image Segmentation with Continual Adaptation}
\paragraph{Retraining on annotations} Since AC only modifies the network’s inputs and not its parameters, the information  provided  by  the  annotations  does  not  improve  the  predictions  globally in the image. Inspired by WTP~\cite{bearman2016s}, we propose to bypass this locality constraint by retraining the network with a few back-propagation cycles per annotation.
Hence, we use the annotations as a sparse ground-truth to interactively retrain the network using a cross entropy loss on these annotated pixels. We note \textbf{f} to represent the neural network parameterized by $\mathbf{\theta}$ and \textbf{x} its inputs.

\paragraph{Regularization} As only a few pixels are annotated among the millions that usually compose a remote sensing image, the ground-truth maps resulting from the interactions are extremely sparse. In order to deal with this problem and avoid over-fitting, we follow ideas from~\cite{kontogianni2020continuous,caye2019guided} by using the initial prediction $\mathbf{p_0} = f(\mathbf{x}, \theta_0)$ for regularization. Precisely we add a $L1$-loss term using the original prediction as reference in order to prevent the model from making a prediction too different from the initial one. Therefore, our loss during the interactive learning process is defined as follows:
\begin{equation}
\label{eq:loss}
\begin{aligned}
\mathcal{L}(\mathbf{x}, \mathbf{c}, \mathbf{p_0}; \mathbf{\theta}) &=  \frac{\mathbf{1}_{[\mathbf{c\ne-1}]}}{\|\mathbf{1}_{[\mathbf{c\ne-1}]}\|_1}\left\{-\sum\limits_{i=1}^{N}\mathbf{c}_i\log\left(\mathbf{f}_i(\mathbf{x};\mathbf{\theta})\right)\right\}\\&+\lambda\|\mathbf{f}(\mathbf{x}; \mathbf{\theta})-\mathbf{p_0}\|_1
\end{aligned}
\end{equation}

where \textbf{1} represents the indicator function and \textbf{c} the sparse annotated pixels. In details, \textbf{c} takes its values in \{-1, 0, 1\}. For the pixels annotated as belonging to class $i$, $\mathbf{c}_i=1$ and $\mathbf{c}_j=0$ for all $j\neq i$. For the unannotated pixels, $\mathbf{c}_i=-1$ for all $i$ in $\{1,\dots,N\}$. $\|\mathbf{1}_{[\mathbf{c\ne-1}]}\|_1$ weights the loss with respect to the number of annotated pixels. Finally, the positive parameter $\lambda$ balances the influence of user annotations with respect to  the recall towards the initial prediction. Its tuning will be considered in Section~\ref{sec:ablation}.

The two last mechanisms enable the continual learning potential of DISCA and avoid catastrophic forgetting. During the interactive training phase, the AC mechanism is randomly disabled: the annotations are then removed from the inputs. This avoids over-fitting on the annotation channels.

% \hl{FOLLOWING PARAGRAPH TO REMOVE ? (Raise an error with the gaston template) BLS: I would say so, what is before is well structured and clear }

% To summarize, the DISCA module leverages three mechanisms:
% \begin{enumerate}
%     \item \gaston{DISIR}{AC}: The annotations modify the neural network inputs.
%     \item WTP: It uses the annotations as a ground-truth for interactive retraining.
%     \item Finally, a regularization term based on initial predictions is crucial to complement the cross-entropy loss during retraining.
% \end{enumerate}
% The two last mechanisms enable the continual learning potential of DISCA and avoid catastrophic forgetting.

% \subsection{Methodology}
\subsection{Active learning component}
% \label{sec: uncertainty}
% \gaston{So far we have presented DISIR and DISCA algorithms that can both be used to refine segmentation proposals made by the network. However, since remote sensing images can be extremely large, it can be tedious for an operator to review them entirely.  To address this issue, we propose to integrate active learning within our framework to swiftly guide the agent towards queries representing the most meaningful areas of the image to annotate.  To this purpose, }{

Since remote sensing images can be extremely large,  DIAL also incorporates an active learning strategy to swiftly guide the agent towards queries representing the most meaningful areas of the image to annotate. With this aim, we compare different state-of-the-art acquisition functions which estimate the algorithm uncertainty to find the most suited to our usage scenario and interaction set-ups described below.

\subsubsection{Formalization}

To formalize the problem,  we note \textbf{f} to represent the neural network parameterized by $\mathbf{\theta}$, \textbf{x} an input image, \textbf{y} its associated label map, \textbf{a} the user annotations and \textbf{g} the annotation encoding function. Our goal is then to find the optimal annotations $\mathbf{a^\star }$ minimizing the following problem: % such as: % $\mathbf{(a^\star ,g^\star )}$

\begin{equation}
\label{eq:overall} % 
% \left\{ 
\begin{aligned}%[l,l]
&\mathbf{a^\star} = \argmin_{\mathbf{a}}\sum\limits_{j\in I}\left( 1-\delta^{u^j}_{\mathbf{y}^j}\right)\\ %\textit{ with }
&\text{with }u^j = \argmax_{c\in\llbracket0,N\rrbracket }\mathbf{f^{\textit{j}}_{\theta\textit{,c}}}\left(\mathbf{x}\oplus\mathbf{g(a, x, f_\theta)}\right)
\end{aligned}
% \right.
\end{equation}

where $\oplus$ represents the concatenation operation, $\delta$ the Kronecker operator, $N$ the cardinal of the label space and $I$ the pixels set. The problem values range from 0 when all pixels are well classified to $card(I)$ when all pixels are misclassified.

\subsubsection{Methodology}

We propose the following query strategy to benefit from DIAL on a given image. The image is divided into a grid of N patches. The patches are annotated consecutively but the order in which they are annotated depends on the uncertainty measure. We have also studied a pixel-based query strategy in Appendix~\ref{sec:pix_based}. 

% In the case of the simulated agent, the clicks are sampled in the patches in the largest erroneous areas using a comparison between the ground-truth and the prediction.
\subsubsection{Acquisition functions} % Uncertainty measures

We now present the different acquisition functions that we compare to guide the agent.

\paragraph{Entropy}
We compute the entropy per pixel at the softmax output: $\mathcal{U} = -\sum_c y_c\times \log (f_c(x;\theta))$. As showed by~\cite{hendrycks2016baseline}, even though the softmax probabilities of a neural network are poorly calibrated, they can still provide a strong baseline to guide the user.
\paragraph{MC Dropout}
MC Dropout~\cite{gal2016dropout} introduces stochasticity in the prediction by enabling dropout regularization at inference time. This allows to obtain an implicit model ensembling. In practice, we add dropout layers in the neural network architecture and then make multiple forward passes through the network to create as many softmax vectors. We then compute the variance of these predictions to measure their disagreement and use it as the uncertainty measure.
\paragraph{ConfidNet}
As proposed by~\cite{corbiere2019addressing}, we train a small auxiliary network to learn to estimate the confidence value of the downstream network using its last layers as inputs.  It is constituted of one transposed convolutional layer and four $3\times 3$ convolutional layers of respectively 32, 120, 64, 32 and 1 output layers. A final sigmoid layer provides the confidence score.
\paragraph{ODIN}
Following~\cite{liang2017principled} which primarily developed this method for outlier detection, we slightly disturb the image input with an adversarial-like attack aiming to enforce the predicted probabilities of the softmax output towards the predicted classes and add a temperature term in the softmax layer. Then, the adversarial examples are feed-forwarded into the network and we use the softmax output maximum class probability as a confidence measure. Formally, we disturb the input with the following perturbation $\mathbf{x} = \mathbf{x} + \varepsilon\Delta_\mathbf{x}\mathcal{L}(f_\theta(x),\hat{y})$ where $\mathcal{L}$ represents the cross-entropy loss, $f_\theta(x)$ the predicted probabilities from the softmax output and $\hat{y}$ the predicted class.

\subsubsection{Computational cost}
Therefore, these approaches have different inference costs inherent to their underlying structure. Indeed, entropy is virtually cost-free since it computes a simple operation directly on the neural network output. On the contrary, MC Dropout is particularly expensive since it requires computing multiple predictions. Despite the extra prediction, ConfidNet is only slightly more expensive than entropy thanks to the small size of the auxiliary network. Finally, ODIN falls between ConfidNet and MC Dropout due to the creation and inference of the adversarial sample.

%% file: sectionsv3/experiments.tex
\section{Experiments} % DISIR \& DISCA experiments

% \gaston{}{In this experiment section, we first assess on three datasets the relevance of AC and DISCA in the refinement and the domain adaptation use-cases, as defined in Section~\ref{intro}. We then show the relevance of the active learning strategy to guide the agent to annotate throughout the image.}

\subsection{Experimental setup}
\label{experimental_setup}
\input{codes/figures/visual_examples}

\subsubsection{The datasets}
We experiment on three semantic segmentation remote sensing datasets: the \textbf{INRIA Aerial Image Labelling dataset}~\cite{maggiori2017dataset} composed of two classes (\textit{buildings} and \textit{not buildings}) and covering more than 800 $\textrm{km}^2$ in  different cities at a 30 cm resolution, the \textbf{Aerial Imagery for Roof Segmentation (AIRS) dataset}~\cite{chen2018aerial} composed of the same two classes and covering  457 $\textrm{km}^2$ in New-Zealand at a 7.5 cm resolution and the \textbf{ISPRS Potsdam dataset}~\cite{rottensteiner2012isprs} composed of 6 classes (\textit{impervious surface, buildings, low vegetation, tree, car} and \textit{clutter}) covering 3~$\textrm{km}^2$ on Potsdam at a 5 cm resolution. The initial training sets are divided into a smaller training set and a validation set with a ratio 80\%-20\%. This allows to synthesize the annotations required to automatically evaluate the framework. 
The images are tiled into patches of size $512\times 512$ with an overlap of size 128 to be processed.
\color{black}
\subsubsection{The hyper-parameters}
We use a neural network based on the LinkNet~\cite{chaurasia2017linknet} architecture but our approach is agnostic from the neural network backbone.

Except in the annotation encoding study, the annotations are encoded into the neural network channels inputs using distance transform. 

For DISCA, during the interactive learning phase, we optimize the weights using 10 stochastic gradient descent passes with a learning rate of $2e^{-6}$ and minimize the loss defined in Eq.~\ref{eq:loss} with the regularization parameter $\lambda$ set to 1.

% \gaston{To automatically evaluate the refinement performances on the validation sets, we sample the annotations from the ground-truth maps in the wrongly predicted areas and measure the Intersection over Union (IoU) evolution.}{}

\subsubsection{Active learning set-up}

For ODIN, we set the perturbation parameter $\varepsilon$ to $\frac{1}{255}$ and the temperature term to 100. 

For MC Dropout, we add a dropout layer between each encoder and decoder block of our architecture, set the dropout rate to 0.1 and compute the variance over 5 different inferences.

The ConfidNet auxiliary network is trained for 10 epochs per dataset with Adam optimizer. 

To automatically evaluate the active learning component, we split the test images into $512\times 512$ patches, sample one annotation per patch and then make a new prediction on this patch using AC-only and DISCA. With DISCA, we retrain the network sequentially with each patch. We study whether the annotation order can be optimized. The annotations are generated inside the spatially-largest mistakes of the patches. We compute the uncertainty globally in the images and then compute an uncertainty score per patch by averaging the uncertainty across all the pixels of the patch.
We compare the uncertainty-ordered sequences to a randomly-drawn one that constitutes the baseline.

% \begin{minipage}{.49\linewidth}
\input{codes/tables/dial_budget}
% \end{minipage}
% \begin{minipage}{.49\linewidth}
\input{codes/tables/times}
% \end{minipage}

\input{codes/figures/uncertainty_measures}
\input{codes/figures/uncertainty_measures_disca}
\input{codes/tables/times_uncertain}

\subsection{Performances and understanding of DIAL mechanisms}
%\subsection{Performances of AC, DISCA \& AL}

As we can observe on Figure~\ref{fig:visual_examples} and on Table~\ref{tab:dial_budget} where a 50 clicks budget has been set, both AC and DISCA successfully enhance the outputs initially proposed by the neural network. DISCA reaches better improvement than AC-only on AIRS and ISPRS: in Table~\ref{tab:dial_budget}, AC's mean gain with active learning is of 2.2\% while DISCA's one is of 2.5\%. Visually, this translates into correction of areas as a whole in a single click, like the buildings or the parking lot of Figure~\ref{fig:visual_examples}. The slight superiority of not retraining the network with DISCA on INRIA is probably due to a better noise robustness. Indeed, INRIA labeling is based on land register and is thus not as conforms with the signal than AIRS or ISPRS. The overall superiority of DISCA has to be also moderated by the inference time of the two algorithms. Indeed, as shown in Table~\ref{tab:times}, DISCA is more than $10\times$ slower than AC-only due to its retraining component.
%%AC over DISCA 

\subsubsection{Active learning with AC}

As we can see on Figure~\ref{fig:uncertainty_measures}, the random order leads to an improvement linear w.r.t. number of processed patches. All active learning schemes speed up the gain in performances by targeting the more uncertain areas. This is particularly noticeable on the AIRS dataset where 50 annotations are enough to reach 75\% of the final improvement. This behavior is probably due to the dataset itself. Indeed, since it covers a lot of rural areas, many images only contain few buildings and the uncertainty measures then allow to quickly show the areas of interest to the user.

Regarding the different uncertainty measures, ODIN is consistently the worst one. Indeed, it is only slightly better than the random order and, contrary to the other methods, its performance is almost linear on the AIRS dataset. This behavior might be explained by the method original purpose. Indeed, while the other methods aim to estimate the model uncertainty, ODIN aims to detect outliers. Though these tasks are related, it appears here that model errors are not due to this type of issue in the image area. Moreover, Table~\ref{tab:times_uncertain} shows that ODIN and MCDropout considerably slow the prediction process (resp. by factors 1.5 and 2) compared to entropy (factor 1)  and ConfidNet (factor 1.2).

ConfidNet and entropy consistently obtain the best performances, with a slight advantage for the former on AIRS and the domain adaptation use-case. However, ConfidNet is also a bit slower and less flexible since it requires to train an additional network for each dataset. Eventually, entropy offers an excellent trade-off between high accuracy performances and fast computation, as it is only slightly slower than a random pick.

\subsubsection{Active learning with DISCA}
Since DISCA slightly modifies the neural network parameters, we recompute the entire prediction and uncertainty after each processed patch. Since MC Dropout and ODIN proved to be relatively slow and less performing with AC, we only compare entropy and ConfidNet in this set-up. As we can observe on Figure~\ref{fig:uncertainty_measures_disca}, results are more complex to interpret than with AC. 

On ISPRS, the different methods are all a bit unstable, which is probably explained by the different improvements for the multiple classes of this dataset. However, both uncertainty methods still perform better than the random strategy and the strategy relying on ConfidNet enables a gain up to 5\% compared to 4\% for the random one. On INRIA, both uncertainty strategies outperform the random one for the first 60 patches but end up being caught up, probably stuck in a local minimum. It is noteworthy that ConfidNet ends up outperforming entropy on these two datasets by a larger margin than with AC. On AIRS and in the domain adaptation situation, the behaviors are similar to the ones obtained with AC, with noticeably higher performances. Indeed, the gain are around 20\% and 5\% with DISCA while they were around 10\% and 4\% with AC respectively on the domain adaptation situation and the AIRS dataset. 

Hence, these results confirm the benefits of a guidance towards relevant patches relying on uncertainty measures. 
ConfidNet is on average the best method to this aim. However, the faster, simpler and only slightly under-performing entropy is a very good alternative for successfully recognizing the most relevant areas to annotate.

% To conclude, active learning complements well the DIAL framework alongside AC and DISCA by guiding the agent towards relevant queries. This is particularly relevant in a limited clicks budget setting, as illustrates Table~\ref{tab:dial_budget} where a 50 clicks budget has been set.

\subsection{Ablation study and comparison with state-of-the-art}
\label{sec:ablation}

\subsubsection{Active learning}
\input{codes/tables/dial_all_tabs}

As shown in the previous section, an active learning patch order leads to better agent annotations than a random patch order with both AC and DISCA. We compare it here to a theoretical upper bound of AC and DISCA: the agent generates each click at the center of the largest spatial error on the whole image, which would be optimal in terms of potential improvement but at the cost of a whole image search. As we can observe in Table~\ref{tab:all_tabs}, this leads, with 10 annotations with AC/DISCA, to an average 1.3/1.7\% improvement over the three datasets against a 1.1/1.5\% improvement with the active learning strategy. However, this slight superiority is mitigated by the complexity to find the annotations. Indeed, in the whole image case, the agent has to browse through $3.6\times 10^7$ pixels for each click in a $6000\times 6000$ image~(complexity: $\mathcal{O}(n_{\mathrm{annots}}*d^2_{\mathrm{image}})$) whilst, in the patch case, it has to browse through $2.6\times 10^5$ pixels in a $512 \times 512$ patch~(complexity: $\mathcal{O}(n_{\mathrm{annots}}*d^2_{\mathrm{patch}})$). Hence, it is 100 times more costly to find the annotation in an entire remote sensing image than in a patch.

\subsubsection{AC \& DISCA}

\input{codes/figures/ablation}
\input{codes/figures/disca_curve}

\input{codes/figures/domain_adapt_examples}
\input{codes/figures/sequential}
To understand the influence of the aspects of the DISCA algorithm, we  analyze separately its different components. AC (ours) adds input layers for annotations~\cite{xu2016deep}, randomly pre-trained on the ground-truth. WTP~\cite{bearman2016s} retrains the model based on a few annotations. DISCA (ours) sums up AC and WTP with regularization with respect to the initial prediction. We also test AC combined with WTP, and WTP combined with regularization. 
To study the importance of the regularization parameter $\lambda$, we test various values $\text{DISCA}_{\lambda = 1}$ and $\text{DISCA}_{\lambda = 10}$. Finally, we also compare our models to FESTA~\cite{hua2021semantic} which trains a neural network on point annotations (as WTP) with a different regularization. 

As shown on Figure~\ref{fig:ablation}, AC and WTP+reg obtain IoU gains around 1\%  for 10 clicks and are beaten by the various flavours of DISCA which almost doubles the gain. This means that the interactive retraining process could be effectively applied to any classically trained neural network but needs to be combined with the AC process to fully exploit the annotations. Moreover, we observe that the regularization is extremely important in DISCA as its absence leads to worse results (AC+WTP curve) than the initial ones (LinkNet curve). A too high $\lambda$ also decreases the benefits brought by DISCA because it then prevents the algorithm to optimally exploit the annotations. Finally, in this framework of incremental learning, WTP~\cite{bearman2016s} and FESTA~\cite{hua2021semantic} also lead to  worse results than the initial ones. These methods were originally designed to train the neural networks from scratch on point annotations. Hence, it explains why they are not optimal in a refinement scenario since they take into account different constraints.

We also compare our approaches with the recent ALCD method~\cite{baetens2019validation} also deployed in the field of remote sensing for cloud segmentation in low resolution~(60 m/pixel) images. To adapt it to our use-case, we run ALCD in a fine-tuning setting on the ISPRS dataset. In practice, we initially pre-train the ALCD random forest on 100000 samples per image from the training set, and then adapt the classifier with the same number of annotations as AC and DISCA. However it leads to very poor performances both before (30\% IoU) and after fine-tuning (30.5\% IoU) compared to AC/DISCA results presented previously. While the absolute results might be due to differences of peculiar implementations of random forest and neural network, the ALCD gain is only $+0.5\%$, which is 2 times less than AC and 3 times less than DISCA.

\subsection{Domain adaptation use-case}

\subsubsection{Performances}

The objective in this domain adaptation use-case is to detect the buildings on the 8 images of the ISPRS validation set. To this purpose, we compare a neural network trained on AIRS under AC and DISCA settings to a control one trained on the ISPRS training set. The ISPRS images are down-sampled using bi-linear interpolation to the AIRS resolution. The neural network's weights are reinitialized between each image. Figure~\ref{fig:disca_curve} shows that a network weakly supervised with DISCA beats AC by a large margin in this scenario. Besides, it can quickly reach high performances (more than 80\% IoU within 20 annotations) and even outperform a fully supervised one with a sufficient amount of annotations. This is visually confirmed on Figure~\ref{fig:domain_adapt_examples}. Indeed, 10 annotations enable the network to well understand the new domain images and thus propose decent segmentation maps. More annotations correct most of the remaining mistakes. %This intends to demonstrate the effectiveness of DISCA in a low initial accuracy scenario.

\subsubsection{Sequential learning}

Moreover, we analyze the generalization of DISCA through a sequence of images in the same domain adaptation scenario. This means that we do not reinitialize the neural network weights between each image. We refer to this set-up as sequential learning, and we learn two insights from it on Figure~\ref{fig:sequential}. First, DISCA does not suffer from catastrophic forgetting here as the algorithm does not diverge even on the last seen images. Second, sequential learning greatly improves the initial performances directly after the first image. Indeed, the initial IoU then approximately increases of 20\%. However, after few annotations, the sequential learning benefits vanish and the performances become similar to the non-sequential set-up.

\subsection{Discussion: when to choose AC-only or DISCA?}
%\subsection{Influence of initial segmentation conditions, or: when to choose AC-only or DISCA?}

To better apprehend the difference between the two methods, we sample 10000 $512\times 512$ crops from each dataset. Then, given one annotation, we compare the difference between AC and DISCA based on two parameters: the spatial size of the corrected mistake and the initial accuracy of the network on the patch. Precisely, the spatial size of the corrected mistake is the size of the error polygon in which the annotation is sampled. Similarly to the initial accuracy, it is obtained with a comparison between the initial predicted map and the ground-truth map. It is intuitively obvious that both AC and DISCA are correlated to these parameters since, if the mistake to correct is small, the overall IoU gain will be smaller than with a larger mistake to correct. However, we think that this comparison can bring valuable insights to choose the appropriate method depending on the situation.

Figure~\ref{fig:correlations2}~compares the two methods with respect to these two criteria. First, both methods seem to work well and can outperform the other one when the mistake area is small and the initial performance is high. We thus recommend to use AC in these situations. Indeed, the locality of AC is no longer a constraint since the error is strongly spatially contained and the relatively long retraining time inherent to DISCA makes it less suitable here. Second, when the initial accuracy is low or the area to correct large, DISCA now clearly tends to perform better than AC, and we thus believe that it should be favored in these situations. Indeed, its spatial globality resulting from its retraining can be fully expressed to correct large mistakes. This outcome shows that DISCA is more relevant to correct deeply flawed segmentation maps than AC.

\input{codes/figures/correlations}

%% file: codes/figures/visual_examples.tex
\begin{figure*}[!htb]
    \newcommand\x{.24}
    \newcommand\xx{\vspace{.1cm}}
    
    \begin{minipage}[t]{\x\linewidth}
        % \centering\epsfig{figure=figures/crop_airs684/gt2.png,width=\linewidth}
        
        % \xx
        
        \centering\epsfig{figure=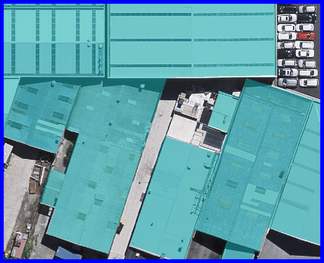,width=\linewidth}
        
        \xx
        
        \centering\epsfig{figure=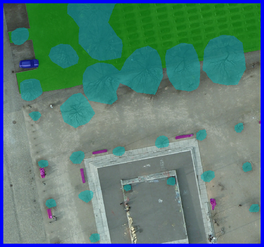,width=\linewidth}
        
        \xx
        
        \centering\epsfig{figure=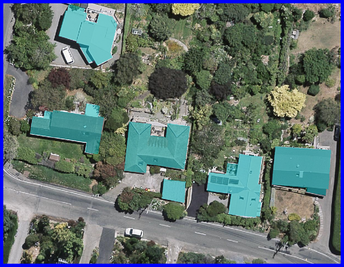,width=\linewidth}

        \xx
        
        \centering\epsfig{figure=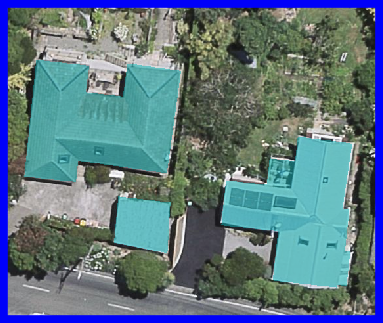,width=\linewidth}
        
        Ground-truth
    \end{minipage}
    % \begin{minipage}[t]{\x\linewidth}
    % \centering\epsfig{figure=figures/crop_airs684/pred.png,width=\linewidth}
    %
    % \xx
    %
    % \centering\epsfig{figure=figures/crop_potsdam513/pred.png,width=\linewidth}
    % 
    % \xx
    % 
    % \centering\epsfig{figure=figures/crop_airs975/pred.png,width=\linewidth}
    % 
    % Initial output
    % \end{minipage}
    \begin{minipage}[t]{\x\linewidth}
        % \centering\epsfig{figure=figures/crop_airs684/annots.png,width=\linewidth}
        
        % \xx
        
        \centering\epsfig{figure=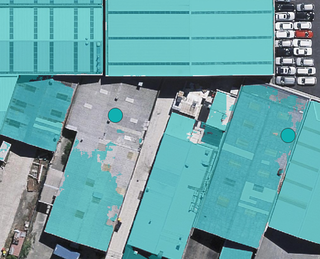,width=\linewidth}
        
        \xx
        
        \centering\epsfig{figure=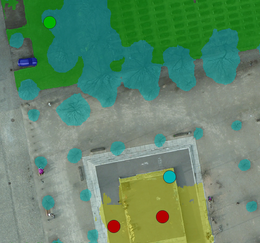,width=\linewidth}
        
        \xx
        
        \centering\epsfig{figure=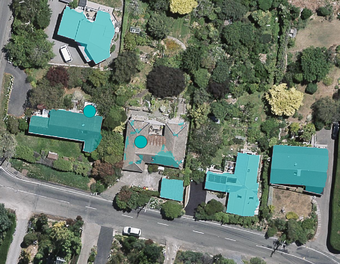,width=\linewidth}

        \xx
        
        \centering\epsfig{figure=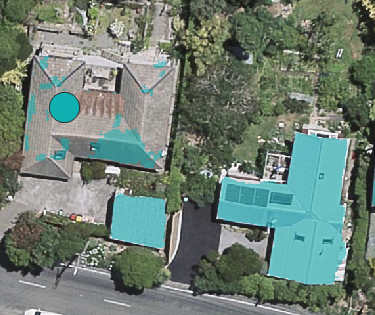,width=\linewidth}
        
        Initial output with annotations
    \end{minipage}
    \begin{minipage}[t]{\x\linewidth}
        % \centering\epsfig{figure=figures/crop_airs684/disir.png,width=\linewidth}
        
        % \xx
        
        \centering\epsfig{figure=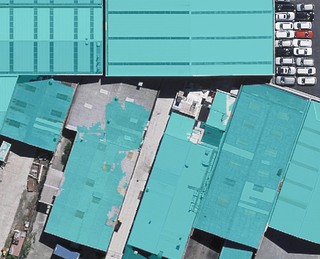,width=\linewidth}
        
        \xx
        
        \centering\epsfig{figure=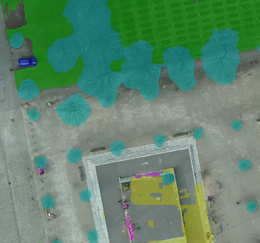,width=\linewidth}
        
        \xx
        
        \centering\epsfig{figure=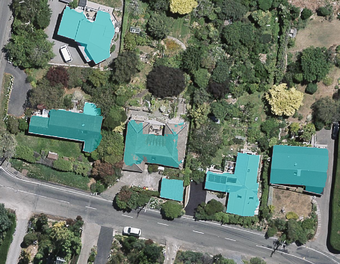,width=\linewidth}

        \xx
        
        \centering\epsfig{figure=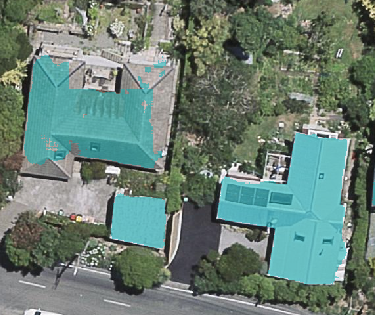,width=\linewidth}
        
        AC output
    \end{minipage}
    \begin{minipage}[t]{\x\linewidth}
        % \centering\epsfig{figure=figures/crop_airs684/disca.png,width=\linewidth}
        
        % \xx
        
        \centering\epsfig{figure=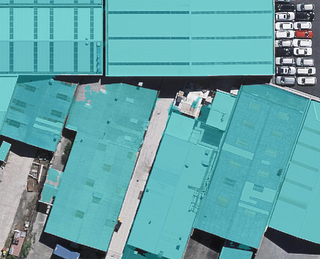,width=\linewidth}
        
        \xx
        
        \centering\epsfig{figure=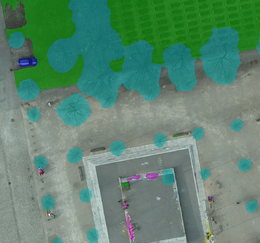,width=\linewidth}
        
        \xx
        
        \centering\epsfig{figure=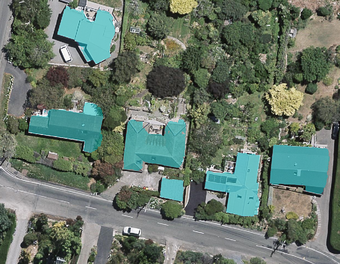,width=\linewidth}

        \xx
        
        \centering\epsfig{figure=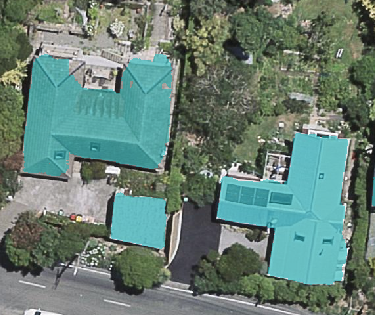,width=\linewidth}    
        
        DISCA output
    \end{minipage}
    	\caption{Visual comparison of the two approaches on examples from AIRS (rows 1, 3 \& 4) \& ISPRS (row 2). Row 4 is a zoomed version of row 3. In rows 1, 3 \& 4, building labels and predictions are in cyan.  In row 2, impervious surface labels are transparent and the associated annotations are in red, buildings are in yellow, low vegetation in green, high vegetation in cyan and clutter in magenta.}
    % 	\caption{Visual comparison of the two approaches on examples from AIRS (rows 1, 2, 4 \& 5) \& ISPRS (row 3). Row 2 and 5 are zoomed version of respectively rows 1 and 4.}
    \label{fig:visual_examples}
\end{figure*}

%% file: codes/tables/dial_budget.tex
\begin{table}[ht]
\caption{Mean IoU after 50 annotated patches with random and active learning (entropy) orders. For 50 patches on Figures~\ref{fig:uncertainty_measures}~\&~\ref{fig:uncertainty_measures_disca}, one recovers results from this Table.}
\centering
\begin{tabular}{c|a??cc??cc}
\toprule % \multirow{2}{*}{\textit{Dataset}}
         &\multicolumn{1}{c??}{\textbf{Initial}}  & \multicolumn{2}{c??}{\textbf{Rand. patches}}  & \multicolumn{2}{c}{\textbf{AL patches}} \\
 & \textit{LinkNet}     & \textit{AC}      & \textit{DISCA}      & \textit{AC}          & \textit{DISCA}       \\ 
\midrule
\textit{ISPRS}   & 70.7                &71.8         & 71.3       & 73.1       & \textbf{73.3 }      \\
\textit{INRIA}   & 85.4                &86.3         & 86.2       & \textbf{86.6}       & 86.4       \\
\textit{AIRS}    & 88                & 88.7         & 89.4       & 91.1       & \textbf{92}     \\
\bottomrule
\end{tabular}
\label{tab:dial_budget}
\end{table}

% \begin{table}[ht]

% \centering
% \begin{minipage}{.49\linewidth}
% \caption{Mean IoU after 50 annotated patches with random and active learning (entropy) orders. For 50 patches on Figures~\ref{fig:uncertainty_measures}\&~\ref{fig:uncertainty_measures_disca}, one recovers results from this Table.}

% \begin{tabular}{c|a??cc??cc}

% \toprule % \multirow{2}{*}{\textit{Dataset}}
%          &\multicolumn{1}{c??}{\textbf{Initial}}  & \multicolumn{2}{c??}{\textbf{Rand. patches}}  & \multicolumn{2}{c}{\textbf{AL patches}} \\
%  & \textit{LinkNet}     & \textit{AC}      & \textit{DISCA}      & \textit{AC}          & \textit{DISCA}       \\ 
% \midrule
% \textit{ISPRS}   & 70.7                &71.8         & 71.3       & 73.1       & \textbf{73.3 }      \\
% \textit{INRIA}   & 85.4                &86.3         & 86.2       & \textbf{86.6}       & 86.4       \\
% \textit{AIRS}    & 88                & 88.7         & 89.4       & 91.1       & \textbf{92}     \\
% \bottomrule
% \end{tabular}
% \label{tab:dial_budget}
% \end{minipage}
% \begin{minipage}{.49\linewidth}
%     \caption{Mean inference time on a $512\times 512$ patch}
%     \begin{tabular}{c|a|cc}
%         \toprule
%          & \textit{Initial} & \textit{AC} & \textit{DISCA} \\ \midrule %
%          \textit{time (s)} & 0.01    & \textbf{0.01}  & 0.11 \\ \bottomrule %
%     \end{tabular}
%     \label{tab:times}
% \end{minipage}
% \end{table}

%% file: codes/tables/times.tex
\begin{table}[ht]
    \centering
    \caption{Mean inference time on a $512\times 512$ patch}
    \begin{tabular}{c|a|cc}
        \toprule
         & \textit{Initial} & \textit{AC} & \textit{DISCA} \\ \midrule %
         \textit{time (s)} & 0.01    & \textbf{0.01}  & 0.11 \\ \bottomrule %
    \end{tabular}
    \label{tab:times}
\end{table}

%% file: codes/figures/uncertainty_measures.tex
\begin{figure}[!h]
\newcommand\x{.49}
\centering
\begin{minipage}[t]{1\linewidth}
    \scriptsize
    \begin{minipage}[t]{\x\linewidth}
        \centering\epsfig{figure=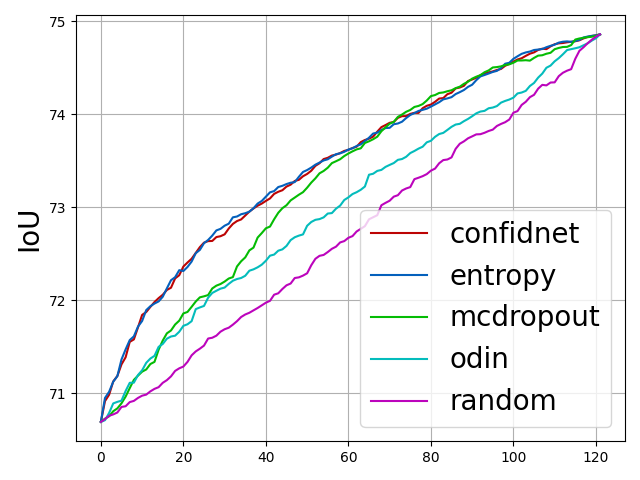,width=\linewidth}
        
        \subcaption{ISPRS}
    \end{minipage}
    \begin{minipage}[t]{\x\linewidth}
        \centering\epsfig{figure=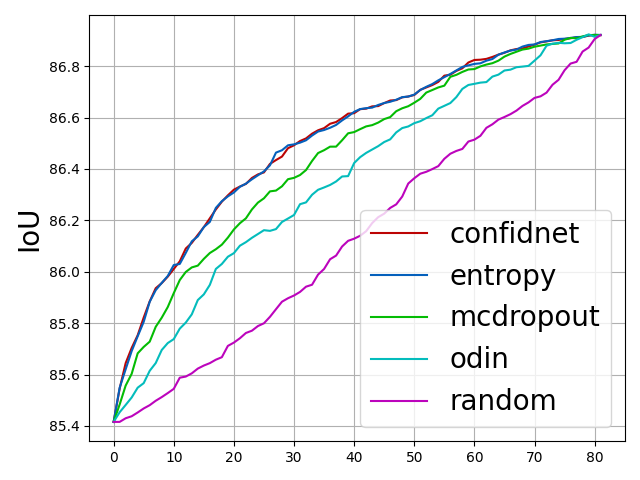,width=\linewidth}
        
        \subcaption{INRIA}
    \end{minipage}
    
    \begin{minipage}[t]{\x\linewidth}
        \centering\epsfig{figure=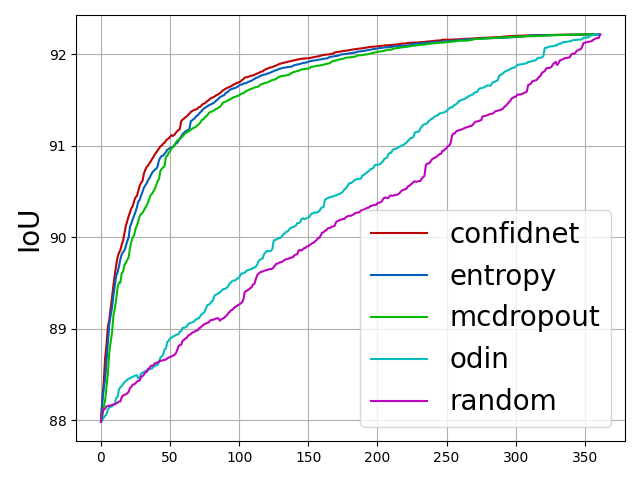,width=\linewidth}
        
        \subcaption{AIRS}
    \end{minipage}
    \begin{minipage}[t]{\x\linewidth}
        \centering\epsfig{figure=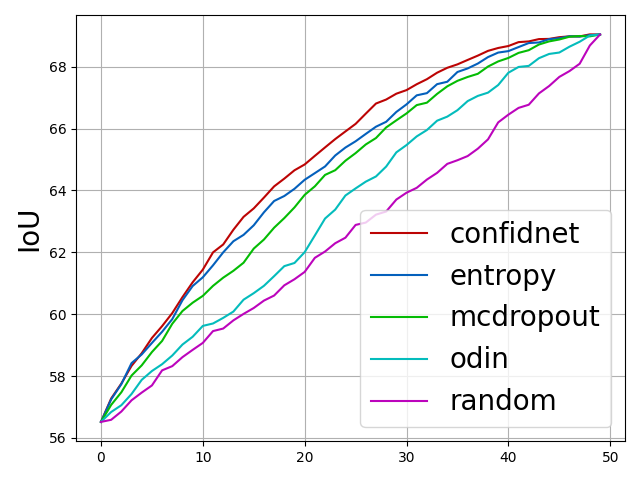,width=\linewidth}
        
        \subcaption{AIRS $\to$ ISPRS}
    \end{minipage}
    	\caption{IoU evolution with respect to the number of annotated patches with AC (one annot. per patch). This compares the different uncertainty measures to select the patch-to-annotate.}
    \label{fig:uncertainty_measures}
\end{minipage}
\end{figure}

%% file: codes/figures/uncertainty_measures_disca.tex
\begin{figure}[!h]
    \newcommand\x{.49}
    \centering
    \begin{minipage}[t]{1\linewidth}
    \scriptsize
    \begin{minipage}[t]{\x\linewidth}
        \centering\epsfig{figure=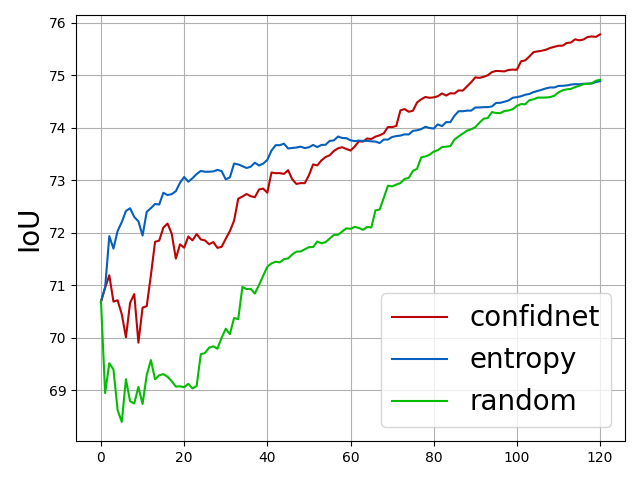,width=\linewidth}
        
        \subcaption{ISPRS}
    \end{minipage}
    \begin{minipage}[t]{\x\linewidth}
        \centering\epsfig{figure=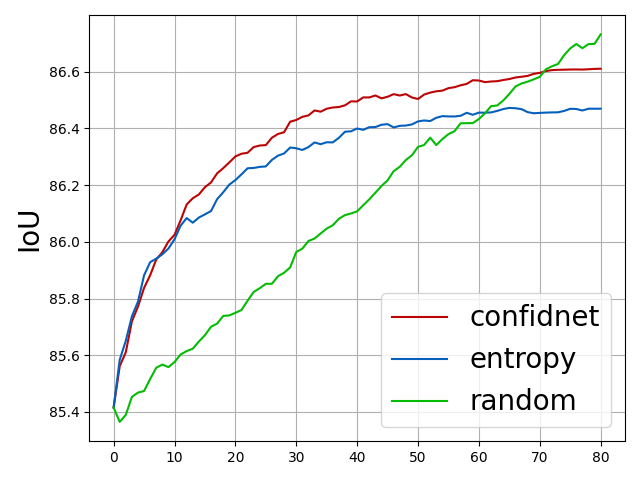,width=\linewidth}
        
        \subcaption{INRIA}
    \end{minipage}
    
    \begin{minipage}[t]{\x\linewidth}
        \centering\epsfig{figure=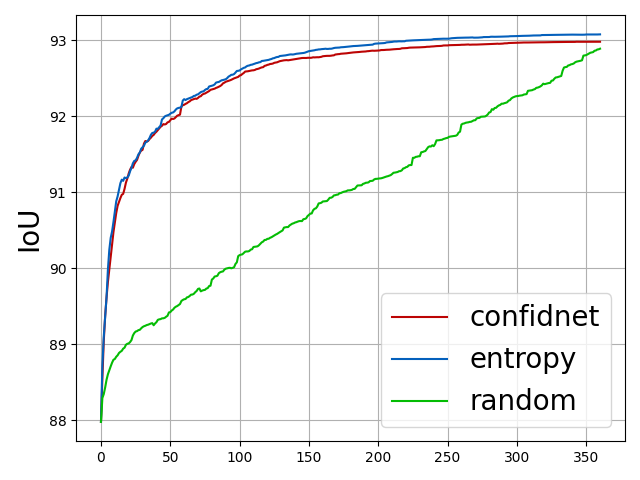,width=\linewidth}
        
        \subcaption{AIRS}
    \end{minipage}
    \begin{minipage}[t]{\x\linewidth}
        \centering\epsfig{figure=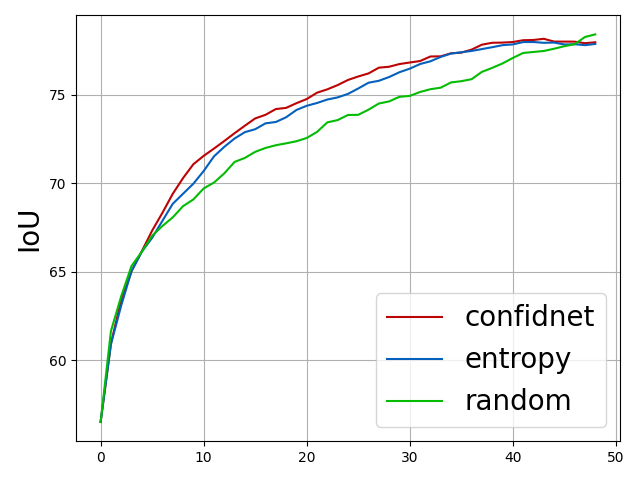,width=\linewidth}
        
        \subcaption{AIRS $\to$ ISPRS}
    \end{minipage}
    	\caption{IoU evolution with respect to the number of annotated patches with DISCA (one annot. per patch). This compares the different uncertainty measures to select the patch-to-annotate.}
    \label{fig:uncertainty_measures_disca}
\end{minipage}
\end{figure}

%% file: codes/tables/times_uncertain.tex
% \begin{table}[]
%   \centering
%     \caption{Mean time (s) for prediction with uncertainty computation on $6000\times 6000$ images}
%     \begin{tabular}{|c|c|c|c|c|}
%         \hline
%          \textit{Random} & \textit{Entropy} & \textit{MC Dropout} & \textit{ODIN} & \textit{ConfidNet}  \\ \hline
%         %  \textbf{0.2}  & 12 & 5.8 & 2.2  \\ \hline
%          9.9 & 10.1  & 22 & 15.8 & 12.6  \\ \hline
%     \end{tabular}
%     \label{tab:times_uncertain}
% \end{table}

\begin{table}[!h]
   \centering
    \caption{Mean time for prediction with uncertainty computation on $6000\times 6000$ images}
    \begin{tabular}{c|a|cccc}
        % \hline
        \toprule
         & \textit{Random} & \textit{Entropy} & \textit{MC Dropout} & \textit{ODIN} & \textit{ConfidNet}  \\ \midrule
        %  \textbf{0.2}  & 12 & 5.8 & 2.2  \\ \hline
         \textit{time (s)} & 9.9 & \textbf{10.1}  & 22 & 15.8 & 12.6  \\ \bottomrule
    \end{tabular}
    \label{tab:times_uncertain}
\end{table}

%% file: codes/tables/dial_all_tabs.tex
\begin{table*}[ht]
\caption{Performances in terms of mean IoU before and after the interactive processes with only 10 annotations per image, w.r.t. corresponding complexity.} %\hl{A modifier pour ajouter une colonne DIAL (qui utilise disca). Éventuellement une deuxieme qui utilise AC ? Ajouter aussi un paragraphe dans la partie set up qui explique comment j'évalue pour AC et disca "naifs". Autre problème: les performances random avec disca en mode dial sont vraiment très mauvaises ...}
\centering
% \resizebox{\columnwidth}{!}{
\begin{tabular}{c|a??cc??cc??cc}
\toprule % \multirow{2}{*}{\textit{Dataset}}
         &\multicolumn{1}{c??}{\textbf{Initial}}  & \multicolumn{2}{c??}{\textbf{Rand. patches}}  & \multicolumn{2}{c??}{\textbf{AL patches}} &  \multicolumn{2}{c}{\textbf{Whole image}} \\
        &\multicolumn{1}{c??}{\textbf{}}  & \multicolumn{2}{c??}{$\mathcal{O}(n_{\mathrm{annots}}*d^2_{\mathrm{patch}})$}  & \multicolumn{2}{c??}{$\mathcal{O}(n_{\mathrm{annots}}*d^2_{\mathrm{patch}})$} &  \multicolumn{2}{c}{ $\mathcal{O}(n_{\mathrm{annots}}*d^2_{\mathrm{image}})$ } \\
 & \textit{LinkNet}     & \textit{AC}      & \textit{DISCA}      & \textit{AC}          & \textit{DISCA}  & \textit{AC}      & \textit{DISCA}        \\ 
\midrule
\textit{ISPRS}   & 70.7                & 71              & 68.7    & 71.8              & 71.9  & 71.7       & \textbf{72.4}      \\
\textit{INRIA}   & 85.4                & 85.5              & 85.5   & 86              & 86  & 86.4       & \textbf{86.5}        \\
\textit{AIRS}    & 88                & 88.2            & 88.8    & 89.6            & \textbf{90.7}   & 89.8       & 90.2     \\
\bottomrule
\end{tabular}
% }
\label{tab:all_tabs}
\end{table*}

%% file: codes/figures/ablation.tex
% \begin{wrapfigure}{r}{0.5\linewidth}
\begin{figure}[!h]
\newcommand\x{1}
\centering
% \vspace{-1.4cm}
    \begin{minipage}[t]{\x\linewidth}
        \centering\epsfig{figure=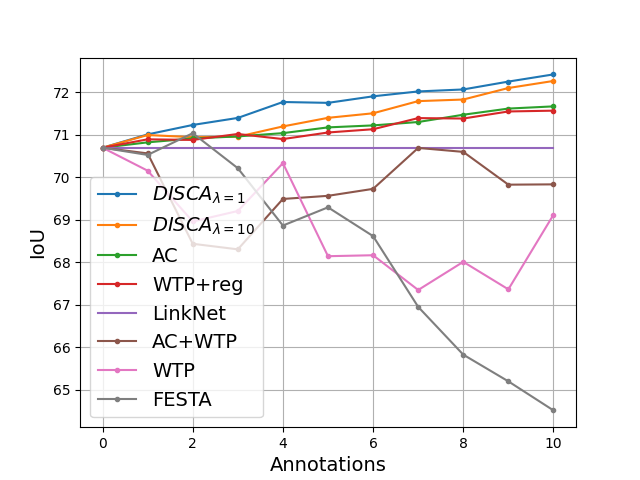,width=0.95\linewidth}
        % \captionsetup{width=.95\linewidth}
        \caption{Ablation study on ISPRS dataset.} % DISCA* corresponds to the DISCA process without using the annotations in the input.
        \label{fig:ablation}
    \end{minipage}
    % \vspace{-.7cm}
    % \centering
% 	\vspace*{-1cm}
\end{figure}

%% file: codes/figures/disca_curve.tex
\begin{figure}[!h]
    \newcommand\x{.49}
    \begin{subfigure}[t]{\x\linewidth}
        \centering\epsfig{figure=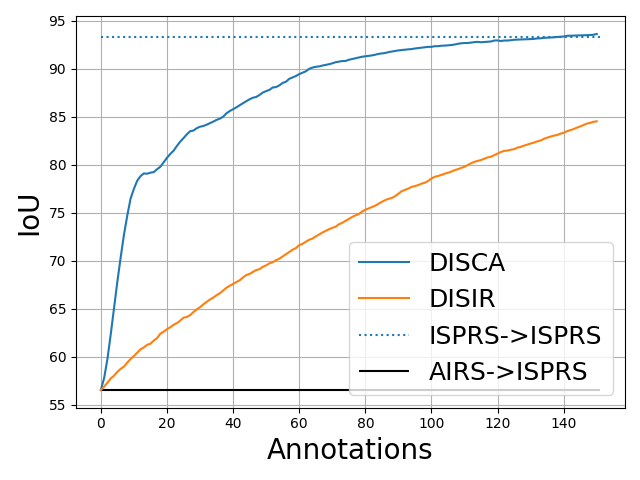,width=\linewidth}
        \captionsetup{width=.95\linewidth}
        \caption{Mean IoU w.r.t the number of annotations.}
    \end{subfigure} \hfill
    \begin{subfigure}[t]{\x\linewidth}
        \centering\epsfig{figure=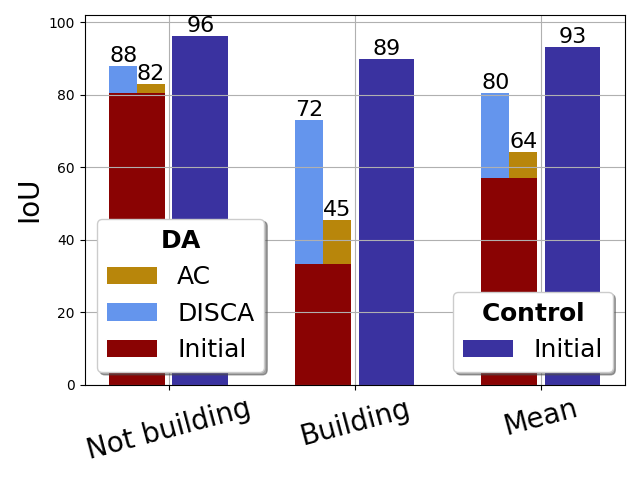,width=\linewidth}
        \captionsetup{width=.95\linewidth}
        \caption{IoU after 20 annotations}
    \end{subfigure}
    \caption{Mean IoU of AC and DISCA for domain adaptation (AIRS$\to$ISPRS).}
    \label{fig:disca_curve}
\end{figure}

%% file: codes/figures/domain_adapt_examples.tex
\begin{figure*}[!h]
    \newcommand\x{.1cm}
    \begin{minipage}[t]{.19\linewidth}
        \centering{\epsfig{figure=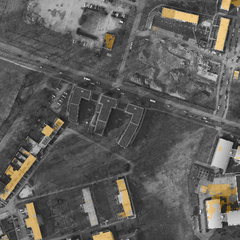,width=\linewidth}}
        
        \vspace{\x}
        
        \centering{\epsfig{figure=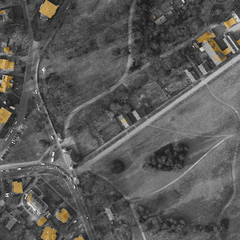,width=\linewidth}}
        
        \vspace{\x}
        
        \centering{\epsfig{figure=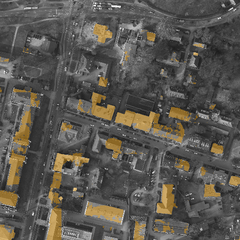,width=\linewidth}}
        
        \vspace{\x}
        
        \centering{\epsfig{figure=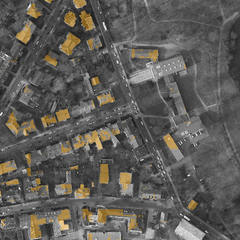,width=\linewidth}}
        
        \vspace{\x}
        
        \centering{\epsfig{figure=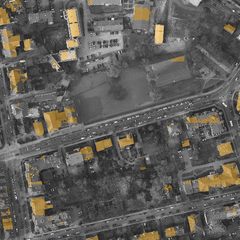,width=\linewidth}}
        
        Initial output
    \end{minipage}\hfill
    \begin{minipage}[t]{.19\linewidth}
        \centering{\epsfig{figure=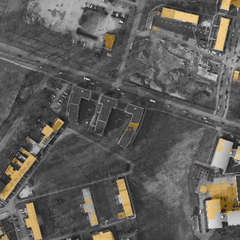,width=\linewidth}}
        
        \vspace{\x}
        
        \centering{\epsfig{figure=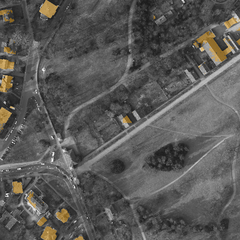,width=\linewidth}}
        
        \vspace{\x}
        
        \centering{\epsfig{figure=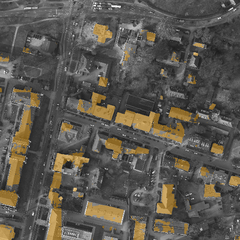,width=\linewidth}}
        
        \vspace{\x}
        
        \centering{\epsfig{figure=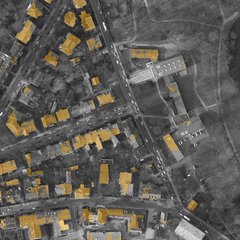,width=\linewidth}}
        
        \vspace{\x}
        
        \centering{\epsfig{figure=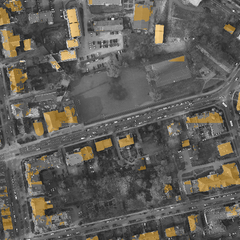,width=\linewidth}}
        
        After 1 annotation
    \end{minipage}\hfill
    \begin{minipage}[t]{.19\linewidth}
        \centering{\epsfig{figure=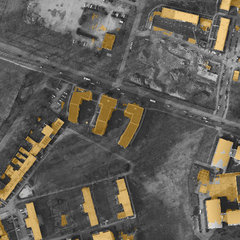,width=\linewidth}}
        
        \vspace{\x}
        
        \centering{\epsfig{figure=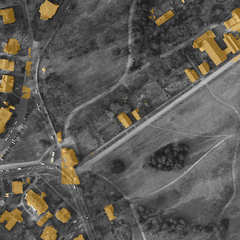,width=\linewidth}}
        
        \vspace{\x}
        
        \centering{\epsfig{figure=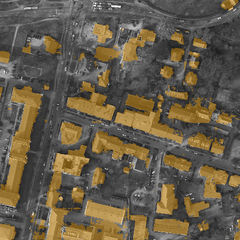,width=\linewidth}}
        
        \vspace{\x}
        
        \centering{\epsfig{figure=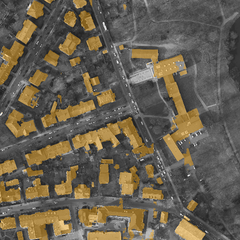,width=\linewidth}}
        
        \vspace{\x}
        
        \centering{\epsfig{figure=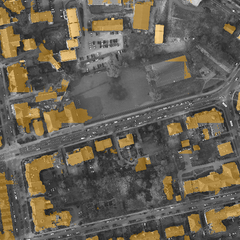,width=\linewidth}}
        
        After 10 annotations
    \end{minipage}\hfill
    \begin{minipage}[t]{.19\linewidth}
        \centering{\epsfig{figure=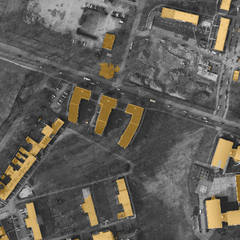,width=\linewidth}}
        
        \vspace{\x}
        
        \centering{\epsfig{figure=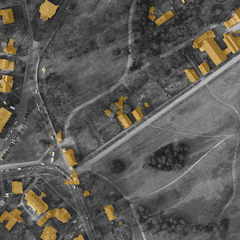,width=\linewidth}}
        
        \vspace{\x}
        
        \centering{\epsfig{figure=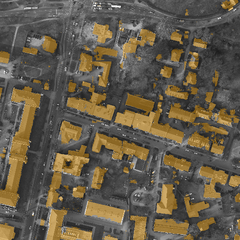,width=\linewidth}}
        
        \vspace{\x}
        
        \centering{\epsfig{figure=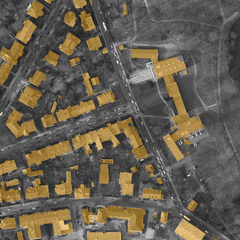,width=\linewidth}}
        
        \vspace{\x}
        
        \centering{\epsfig{figure=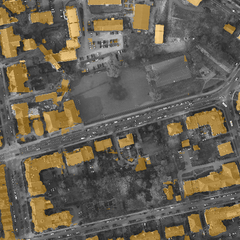,width=\linewidth}}
        
        After 50 annotations
    \end{minipage}
    \begin{minipage}[t]{.19\linewidth}
        \centering{\epsfig{figure=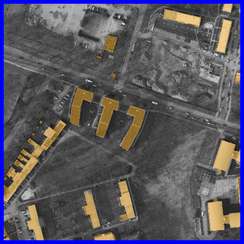,width=\linewidth}}
        
         \vspace{\x}
        
        \centering{\epsfig{figure=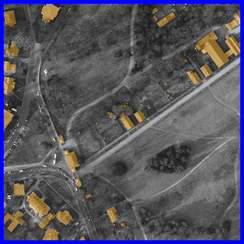,width=\linewidth}}
        
        \vspace{\x}
        
        \centering{\epsfig{figure=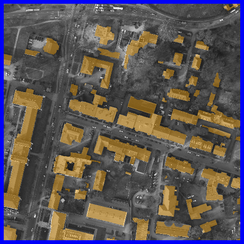,width=\linewidth}}
        
        \vspace{\x}
        
        \centering{\epsfig{figure=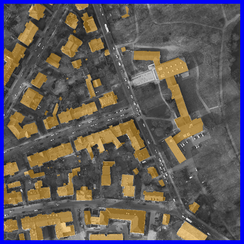,width=\linewidth}}
        
        \vspace{\x}
        
        \centering{\epsfig{figure=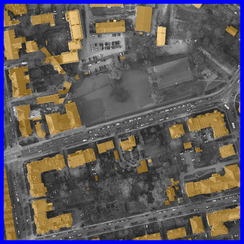,width=\linewidth}}
        
        Ground-truth
    \end{minipage}\hfill
    \caption{Domain adaptation (AIRS $\to$ ISPRS) visual examples}
    \label{fig:domain_adapt_examples}
\end{figure*}

%% file: codes/figures/sequential.tex
\begin{figure}[!h]
    \newcommand\x{.49}
    \begin{subfigure}[t]{\x\linewidth}
        \centering\epsfig{figure=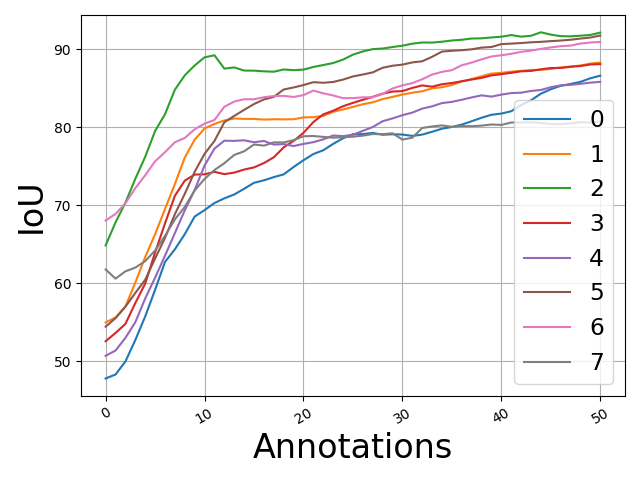,width=\linewidth}
        \captionsetup{width=1\linewidth}
        \caption{Weights reinitialized between each image.}
    \end{subfigure} \hfill
    \begin{subfigure}[t]{\x\linewidth}
        \centering\epsfig{figure=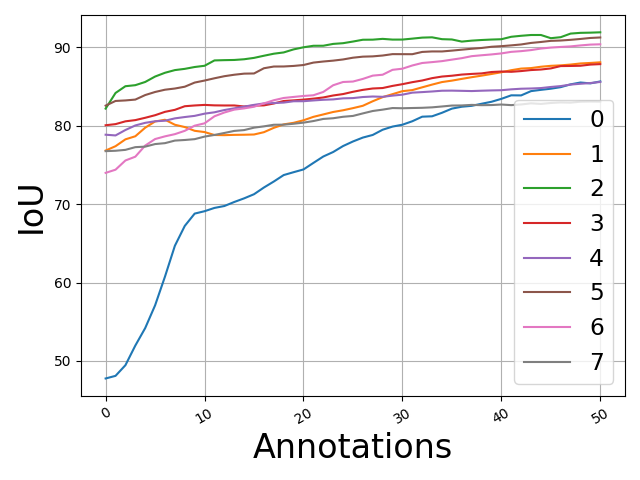,width=\linewidth}
        \captionsetup{width=1\linewidth}
        \caption{Weights updated between each image.}
    \end{subfigure}
    % 
    % \centering
    % \begin{minipage}[t]{\x\linewidth}
    %     \centering\epsfig{figure=figures/sequential_annots_learn.png,width=\linewidth}
    % \end{minipage}
    % \vspace{-.4cm}
    \caption{Sequential learning study with DISCA in a transfer scenario. The legend corresponds to the order in which the algorithm processes the images.}
    \label{fig:sequential}
\end{figure}

%% file: codes/figures/correlations.tex
% \begin{figure*}[h]
%     \begin{minipage}[t]{1\linewidth}
%         \centering\epsfig{figure=figures/statistics/correlations.png,width=\linewidth}
%     \end{minipage}
% 	\caption{Influence of the mistake area and initial accuracy on the performances}
%     \label{fig:correlations}
% \end{figure*}

% \begin{figure}[h]
%     \newcommand\x{.49}
%     \scriptsize
%     \begin{subfigure}[t]{\x\linewidth}
%         \centering\epsfig{figure=figures/statistics/area_gain.png,width=\linewidth}
%         % \captionsetup{width=.95\linewidth}
%         % \caption{IoU gain with DISIR with respect to the corrected mistake area}
%         % \caption{}
%     \end{subfigure}
%     \begin{subfigure}[t]{\x\linewidth}
%         \centering\epsfig{figure=figures/statistics/acc_gain.png,width=\linewidth}
%         % \captionsetup{width=.95\linewidth}
%         % \caption{IoU gain with DISIR with respect to the initial accuracy}
%         % \caption{}
%     \end{subfigure}
%     \caption{Influence of the mistake area and initial accuracy on DISIR performances (IoU).}
%     % 	\caption{Influence of the spatial size of the corrected mistake and initial accuracy on the performances. Legend "best" designates the best method for the given sample.}
%     \label{fig:correlations}
% \end{figure}

\begin{figure}[!h]
    \newcommand\x{.49}
    \scriptsize
    \begin{subfigure}[t]{\x\linewidth}
        \centering\epsfig{figure=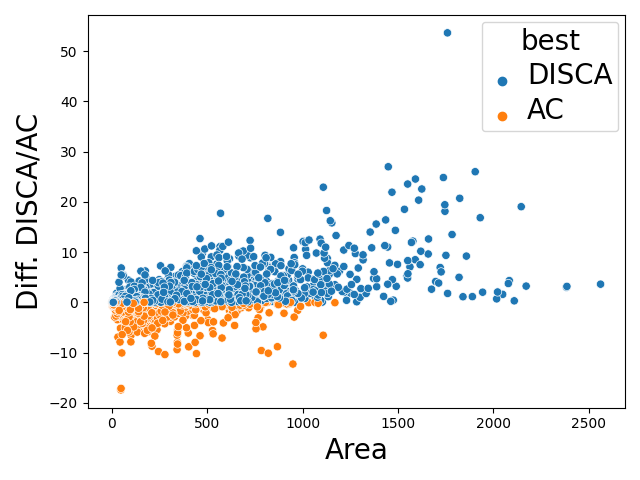,width=\linewidth}
        % \captionsetup{width=.95\linewidth}
        % \caption{IoU gain difference between DISIR and DISCA with respect to the corrected mistake area}
        % \caption{}
    \end{subfigure}
    \begin{subfigure}[t]{\x\linewidth}
        \centering\epsfig{figure=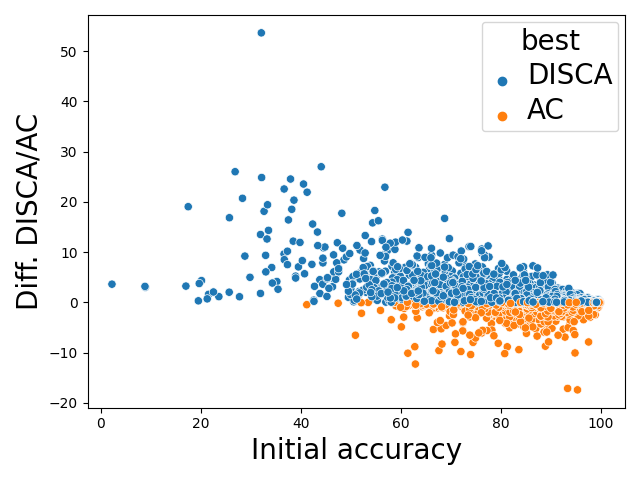,width=\linewidth}
        % \captionsetup{width=.95\linewidth}
        % \caption{IoU gain difference between DISIR and DISCA with respect to the initial accuracy}
        % \caption{}
    \end{subfigure}
    	\caption{Comparison of AC-only and DISCA (IoU) with respect to the spatial size of the corrected mistake and the initial accuracy. Legend "best" designates the best method for the given sample.}
    \label{fig:correlations2}
\end{figure}

%% file: sectionsv3/conclu.tex
\section{Conclusion}
We have presented DIAL, a framework to interactively enhance segmentation maps initially proposed by a neural network. Its core concept relies on interactions between a deep neural network and an agent under the form of clicked annotations. 

% \hl{HERE I REPLACE THE COMMENTED TEXT (STILL DISIR/DISCA Oriented) BY THE SUMMARY PREVIOUSLY CANCELED IN IL PART.} It uses the annotations as a ground-truth for interactive retraining.

First, we have proposed an interactive learning framework which builds on complementary mechanisms. First, Annotations as Channels (AC) modifies the neural network inputs. This approach is fast and local since it does not modify the weights of network. Second, an on-the-fly retraining uses the annotations as a a sparse ground-truth. Finally, a regularization term based on the initial prediction is crucial to complement the cross-entropy loss during retraining and avoids catastrophic forgetting. Since this modifies the weights of the network, the full framework is slower but improves the segmentation maps at a larger scale.

%First, we have proposed two complementary algorithms to efficiently use these annotations. The first one, \gaston{DISIR}{AC}, uses them directly to modify its inputs. This approach is fast and local since it does not modify the weights of network. The second one, DISCA, builds on \gaston{DISIR}{AC} and also uses the annotations as a sparse ground-truth to retrain the neural network on-the-fly. Since this modifies the weights of the network, it is slower but improves the segmentation maps at a larger scale. Through experiments, we have demonstrated the effectiveness of these algorithms and have notably showed that \gaston{DISIR}{AC} can be more convenient to correct spatially small mistakes while DISCA is more tailored for larger mistakes, as it may happen in a domain adaptation use-case.

Finally, we have integrated active learning within our framework to guide the agent interventions towards relevant patch queries. To this purpose, we have compared different state-of-the-art acquisition functions to estimate the neural network uncertainty to finally conclude that entropy is the most suited one thanks to its simplicity and efficiency. Hence, active learning speeds up the use of our interactive segmentation algorithms and is particularly relevant to face budget constraints.

% \gaston{We have identified and proposed two possible scales for the agent guidance. First, patch-based queries lead the agent towards relevant areas to annotate in the images. Second, even though pixel-based queries do not lead to better corrections compared to the annotations inside the spatially large mistakes, they can nonetheless help the agent to spot mistakes and ambiguous zones in the images. Therefore, active learning speeds up the use of our interactive segmentation algorithms.}{

% Indeed, although the optimal annotations seem to be in the middle of the mistake areas, uncertainty measurements can be used to guide the agent to easily spot these mistakes and towards relevant areas to annotate both globally (i.e. by proposing the most relevant patches within an image) and locally (i.e. by indicating the most uncertain pixels). Moreover, we have compared different uncertainty measures from the literature for our use-case and showed that the entropy is simple-yet-efficient in this case.

In the future, we intend to apply DIAL in a class-incremental scenario to make it easily adaptive to new tasks and use-cases. This, along with the promising results shown in domain adaptation, will provide Earth observation scientists and companies with a powerful to re-use, transfer and enhance deep learning models.

% In such context, we have showed that using DISCA on a sequence of images can also be beneficial for the network even though the knowledge gap is filled within a few clicks.

% To get a deeper understanding of our framework, we have analyzed it statistically and this illustrated that the annotations impact is linearly correlated with the size of the mislabelled annotated area. We have also tried to optimize the annotations encoding using the image or the prediction contexts but have only beaten our no-context baseline by an insignificant margin, showing that the network is able to learn to optimally use the annotations information by itself. 

%% file: sectionsv3/supp_material.tex
\section{How to optimize the encoding ?}
\input{codes/figures/encoding}
\input{codes/tables/encodings}

% Now that we have validated the relevance of our algorithms under different scenarios, we aim to optimize the agent guidance and make the best out of our proposed methods. 
% To this aim, 
We investigate here the annotations encoding to analyze its influence on the AC mechanism.

There are many possibilities to encode the annotations in their dedicated channels and they all provide different spatial information. The size of the encoding is the most obvious issue: if it is too small, it might not provide enough information to efficiently fix the initial segmentation but a coarser encoding might provide erroneous information. A popular context-free trade-off used in most interactive segmentation works~\cite{xu2016deep,liew2017regional} is to encode the annotations with Euclidean distance transforms to dilute spatial information. However, due to its context independence, this encoding might be sub-optimal. Ideally, the perfect encoding would be the original ground-truth map but it is obviously impossible to get. Based on this insight, we  study here how to best approximate this ground-truth given the available data: the input image, the annotations and the trained neural network. We define two possible context use besides the no-context one:
\begin{itemize}
    \item Using the input image
    \item Using the initial prediction
\end{itemize}

As encoding baselines, we use small binary (bin.) disks of 1.5 pixels radius and distance transform (DT) applied on 10 pixels radius disks. We build on this DT encoding for the context encodings. To use the input image, we rely on guided filtering~\cite{he2010guided} (GF) in order to preserve the edges in the encoding. To use the initial prediction, we encode the annotations using their connected pixels in the prediction map (C-PM). To estimate the superior boundary theoretically reachable with an encoding from the ground-truth, we also encode the annotations using their connected pixels in the ground-truth map (C-GT). These different methods to encode the annotations are represented in Figure~\ref{fig:encoding}.

However, as shown in Table~\ref{tab:encodings}, the different encoding strategies seem to provide similar information to the network as the gains are in the same order of magnitude. Indeed, they all increase the IoU of around 6\% for 120 annotations on the ISPRS images, even though the binary encoding is slightly lower and confirms the usefulness of DT encoding. The GF encoding obtains the same gain as the DT one, C-PM is lower of 0.1\% and even C-GM is only better of 0.1\%.

These insignificant differences show that the network does not need any contextual guidance to learn nearly optimal information from the annotations using a simple and intuitive encoding such as distance transform. 
%baseline is  strong and is only beaten by a small margin.

\input{codes/figures/uncertainty_gt}

\input{codes/figures/uncertainty_nogt}
\input{codes/figures/uncertainty_examples}
\section{Active learning for pixel-based guidance}
\label{sec:pix_based}
\subsection{Set-up and objectives}
To investigate local guidance, we sample 10000 $512\times 512$ crops from each dataset, make an initial prediction, generate one annotation per sample and then do a new prediction with AC-only and DISCA. Our acquisition function here is entropy. We test two conjectures. First, we want to determine whether highly uncertain pixels among the misclassified ones can lead towards particularly meaningful annotations. Second, we want to figure out whether the uncertainty measurements can help the agent to spot errors at a pixel level.

\subsection{Uncertainty for optimal annotations}

We make the hypothesis that an agent always clicks on a wrongly segmented area, or in other words that he is able to spot the mistakes and correct them. To look for optimal annotations, we compare the following annotations sampling strategies. 
\begin{enumerate}
    \item We sample the annotation randomly in the wrongly segmented area (\textit{random}).
    \item Like in the other experiments, we sample the annotation in the middle of the spatially-largest wrongly-segmented area (\textit{max}).
    \item We threshold the uncertainty map at the ninth quantile to keep only the highest uncertainty values. We then sample the annotation in the intersection of the wrongly-segmented area and this thresholded uncertainty map.
\end{enumerate}

As shown on Figure~\ref{fig:uncertainty_gt}, the uncertainty-based annotations lead to corrections of the same magnitude than the random ones on average.  Moreover, these uncertainty-based annotations clearly don't provide more information to the model than the ones based on \textit{max}. Indeed, the gains of \textit{max} annotations with AC-only are around 6.4\% IoU against respectively 4.5\% and 4.7\% for the random and uncertainty-based ones. 

This corroborates the correlation between the gain and the size of the corrected area previously exhibited and shows that uncertainty does not lead towards more meaningful annotations than the ones contained inside large mistakes. 

\subsection{Uncertainty to spot mistakes}

In order to evaluate if the uncertainty measures can help to spot mistakes at the pixel level, we compare annotations sampled randomly and on the basis of uncertainty measures \textit{without} ground-truth prior knowledge. In other words, we do not coerce the annotations to be sampled in mistake areas. %(which is the way to emulate a smart corrector agent in simulations)

Figure~\ref{fig:uncertainty_nogt} shows that the uncertainty-based annotations lead to better improvements~(3.7\% IoU with AC-only, 4.5\% with DISCA) than the random ones~(1.3\% IoU with AC-only, 1.9\% with DISCA) on average. We can visually confirm these insights on Figure~\ref{fig:uncertainty_examples} where uncertainty measures tend to highlight  wrongly predicted areas. Besides, the highlighted areas which are initially correctly predicted tend to be legitimately questionable such as object contours or road surfaces looking like buildings~(third row).

%% file: codes/figures/encoding.tex
\begin{figure}[!h]
\newcommand\x{.24}
% \begin{minipage}[t]{\x\linewidth}
% \centering\epsfig{figure=figures/guided_filter/imgt.png,width=\linewidth}
% Ground-truth overlaid on input image
% \end{minipage}\hfill
%
\begin{minipage}[t]{\x\linewidth}
\centering\epsfig{figure=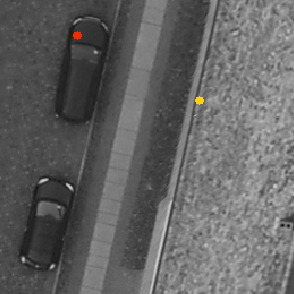,width=\linewidth}
\subcaption{Car (red) and building (yellow) annotations}
\end{minipage}\hfill
\begin{minipage}[t]{\x\linewidth}
\centering\epsfig{figure=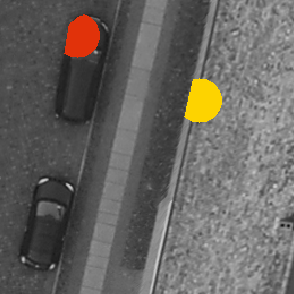,width=\linewidth}
\subcaption{Ground-truth connectivity encoding}
\end{minipage}\hfill
\begin{minipage}[t]{\x\linewidth}
\centering\epsfig{figure=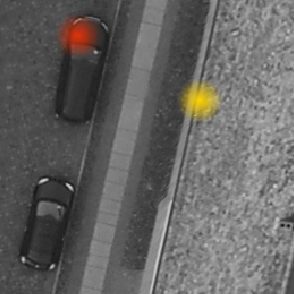,width=\linewidth}
\subcaption{Distance-transform encoding}
\end{minipage}\hfill
\begin{minipage}[t]{\x\linewidth}
\centering\epsfig{figure=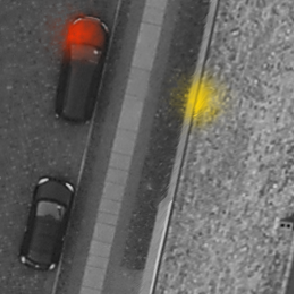,width=\linewidth}
\subcaption{Guided filter encoding}
\end{minipage}\hfill
\caption{Different annotations encodings depending on context uses. Best viewed in color.}
\label{fig:encoding}
\end{figure}

%% file: codes/tables/encodings.tex
\begin{table}[ht]
\centering
\caption{IoU on ISPRS after 120 annotations with AC depending on the encoding.}
    \begin{tabular}{c|ccccc}
    \toprule
               & \textit{Bin.} & \textit{DT} & \textit{C-PM} & \textit{C-GT} (sup) & \textit{GF} \\ \midrule
        \textit{Initial} & 70.7 & 70.7 & 70.7 & 70.7 & 70.8                    \\ 
        \textit{After}  & 76.4                      & 76.6                    & 76.5                                       & \textbf{76.7}                               & 76.7                    \\
        \hline
        \textit{Gain}   & 5.7                       & 5.9                     & 5.8                                        & \textbf{6}                                  & 5.9                     \\ 
        \bottomrule
    \end{tabular}
    \label{tab:encodings}
\end{table}

% \begin{table}[]
% \centering
% \caption{Encoding results on Potsdam dataset after 120 annotations.}
%     \begin{tabular}{|l|l|c|l|l|l|}
%         \hline
%               & \multicolumn{1}{l|}{\textit{Bin.}} & \multicolumn{1}{l|}{\textit{DT}} & \multicolumn{1}{l|}{\textit{C-PM}} & \multicolumn{1}{l|}{\textit{C-GT} (sup)} & \multicolumn{1}{l|}{GF} \\ \hline
%         Before & \multicolumn{4}{c|}{70.7}                                                                                                             & 70.8                    \\ \hline
%         After  & 76.4                      & 76.6                    & 76.5                                       & 76.7                               & 76.7                    \\ \hline
%         Gain   & 5.7                       & 5.9                     & 5.8                                        & 6                                  & 5.9                     \\ \hline
%     \end{tabular}
%     \label{tab:encodings}
% \end{table}

%% file: codes/figures/uncertainty_gt.tex
% \begin{figure}[!h]
% \newcommand\x{.49}
%     \begin{subfigure}[t]{\x\linewidth}
%         \centering\epsfig{figure=figures/uncertainty_local_gt.png,width=\linewidth}
%         \captionsetup{width=.95\linewidth}
%         \caption{Annotations sampled in the mislabelled areas using ground-truth}
%         \label{fig:uncertainty_gt}
%     \end{subfigure}
%     \begin{subfigure}[t]{\x\linewidth}
%         \centering\epsfig{figure=figures/uncertainty_local_nogt.png,width=\linewidth}
%         \captionsetup{width=.95\linewidth}
%         \caption{Annotations sampled without using ground-truth}
%         \label{fig:uncertainty_nogt}
%     \end{subfigure}
%     % \vspace{-1cm}
%     \caption{Study of the relevance of using uncertainty to select the click locally inside a patch with DISIR.}
%     \label{fig:uncertainty}
% \end{figure}

\begin{figure}[!h]
\newcommand\x{.49}
    \begin{subfigure}[t]{\x\linewidth}
        \centering\epsfig{figure=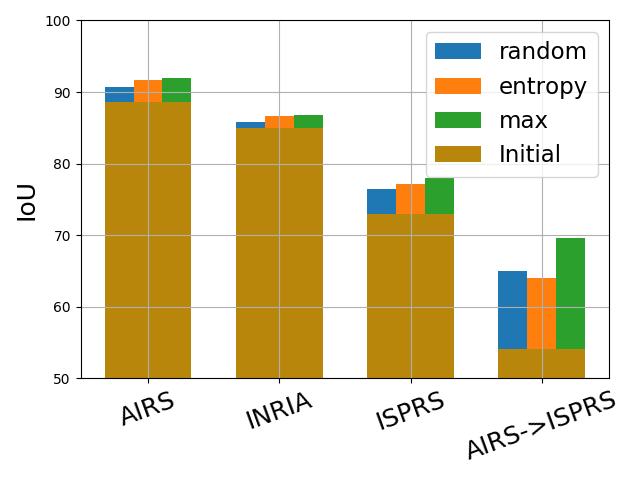,width=\linewidth}
        \captionsetup{width=.95\linewidth}
        \caption{AC-only}
        % \label{fig:uncertainty_gt}
    \end{subfigure}
    \begin{subfigure}[t]{\x\linewidth}
        \centering\epsfig{figure=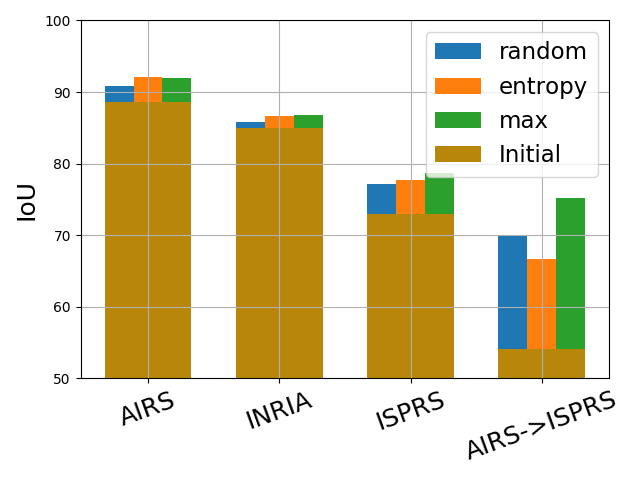,width=\linewidth}
        \captionsetup{width=.95\linewidth}
        \caption{DISCA}
        % \label{fig:uncertainty_nogt}
    \end{subfigure}
    % \vspace{-1cm}
    \caption{Annotations sampled using uncertainty and error knowledge (smart corrector agent)}%Study of the relevance of using uncertainty to select the click locally inside a patch with DISIR.} % Annotations  sampled  in  the  mislabelled  areas  usingground-truth
    \label{fig:uncertainty_gt}
\end{figure}

%% file: codes/figures/uncertainty_nogt.tex
\begin{figure}[!h]
\newcommand\x{.49}
    \begin{subfigure}[t]{\x\linewidth}
        \centering\epsfig{figure=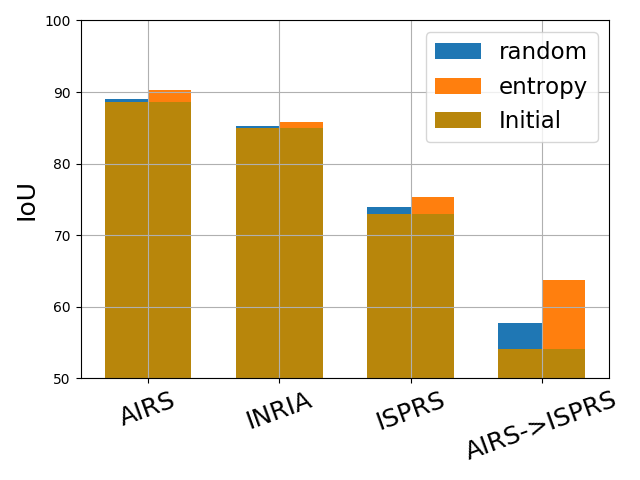,width=\linewidth}
        \captionsetup{width=.95\linewidth}
        \caption{AC-only}
        % \label{fig:uncertainty_gt}
    \end{subfigure}
    \begin{subfigure}[t]{\x\linewidth}
        \centering\epsfig{figure=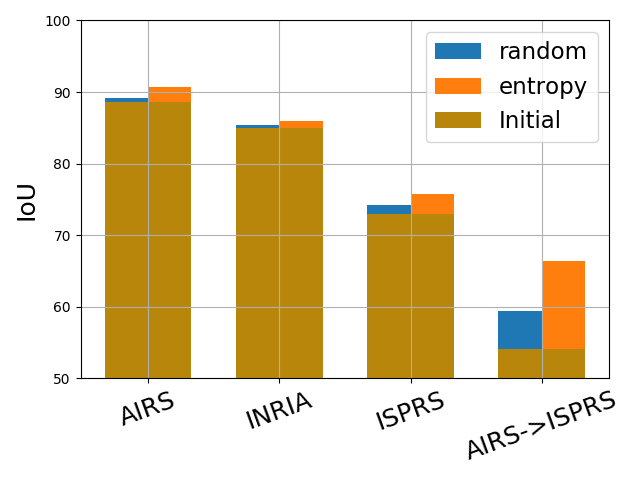,width=\linewidth}
        \captionsetup{width=.95\linewidth}
        \caption{DISCA}
        % \label{fig:uncertainty_nogt}
    \end{subfigure}
    % \vspace{-1cm}
    \caption{Annotations sampled with uncertainty but without error knowledge}
    \label{fig:uncertainty_nogt}
\end{figure}

%% file: codes/figures/uncertainty_examples.tex
\begin{figure*}[!h]
    \newcommand\x{.1cm}

    \begin{subfigure}[t]{.19\linewidth}
        \centering{\epsfig{figure=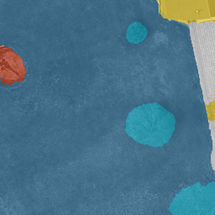,width=\linewidth}}
        
        \vspace{\x}
        
        \centering{\epsfig{figure=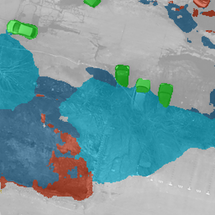,width=\linewidth}}
        
        \vspace{\x}
        
        \centering{\epsfig{figure=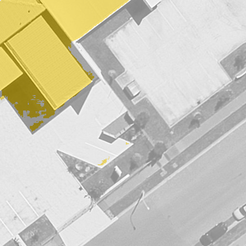,width=\linewidth}}
        
        \vspace{\x}
        
        \centering{\epsfig{figure=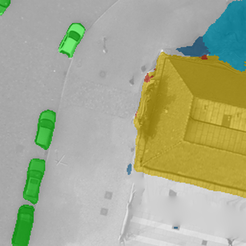,width=\linewidth}}
        
        \vspace{\x}
        
        \centering{\epsfig{figure=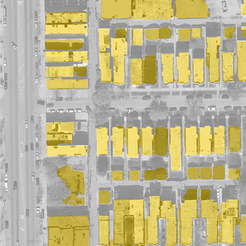,width=\linewidth}}
        
        \vspace{\x}
        
        \centering{\epsfig{figure=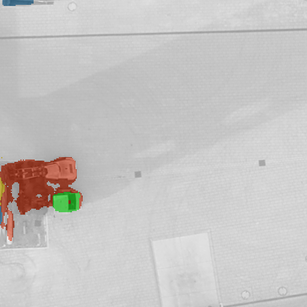,width=\linewidth}}
        
        \captionsetup{width=.95\linewidth}
        Initial output
    \end{subfigure}\hfill
    \begin{subfigure}[t]{.19\linewidth}
        \centering{\epsfig{figure=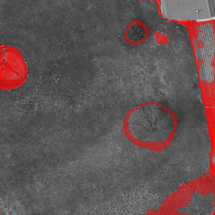,width=\linewidth}}
        
        \vspace{\x}
        
        \centering{\epsfig{figure=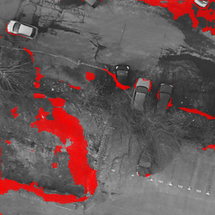,width=\linewidth}}
        
        \vspace{\x}
        
        \centering{\epsfig{figure=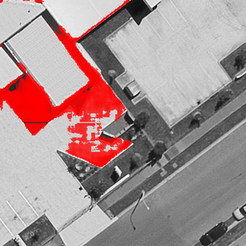,width=\linewidth}}
        
        \vspace{\x}
        
        \centering{\epsfig{figure=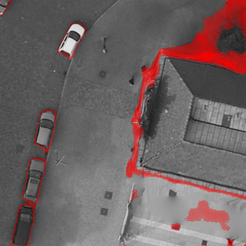,width=\linewidth}}
        
        \vspace{\x}
        
        \centering{\epsfig{figure=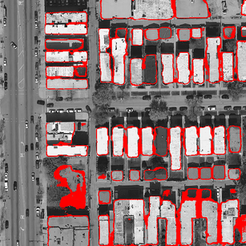,width=\linewidth}}
        
        \vspace{\x}
        
        \centering{\epsfig{figure=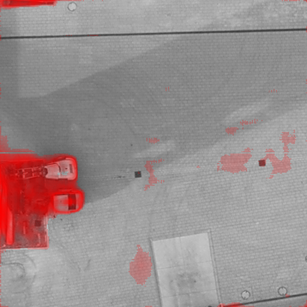,width=\linewidth}}
        
        \captionsetup{width=.95\linewidth}
        Entropy
    \end{subfigure}\hfill
    \begin{subfigure}[t]{.19\linewidth}
        \centering{\epsfig{figure=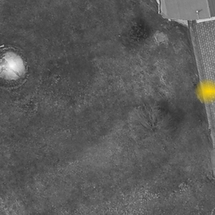,width=\linewidth}}
        
        \vspace{\x}
        
        \centering{\epsfig{figure=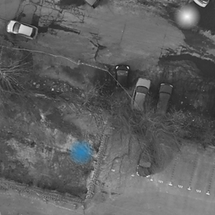,width=\linewidth}}
        
        \vspace{\x}
        
        \centering{\epsfig{figure=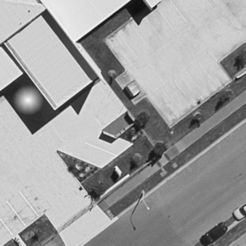,width=\linewidth}}
        
        \vspace{\x}
        
        \centering{\epsfig{figure=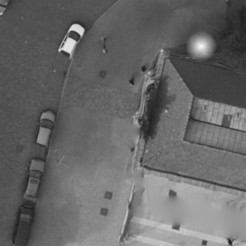,width=\linewidth}}
        
        \vspace{\x}
        
        \centering{\epsfig{figure=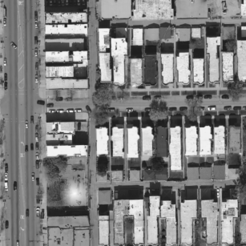,width=\linewidth}}
        
        \vspace{\x}
        
        \centering{\epsfig{figure=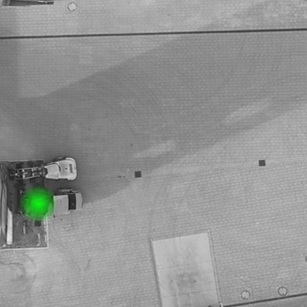,width=\linewidth}}
        
        \captionsetup{width=.95\linewidth}
        Annotation(s)
    \end{subfigure}\hfill
    \begin{subfigure}[t]{.19\linewidth}
        \centering{\epsfig{figure=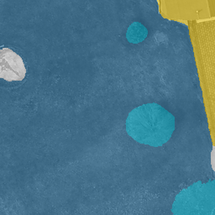,width=\linewidth}}
        
        \vspace{\x}
        
        \centering{\epsfig{figure=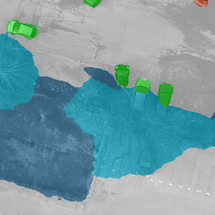,width=\linewidth}}
        
        \vspace{\x}
        
        \centering{\epsfig{figure=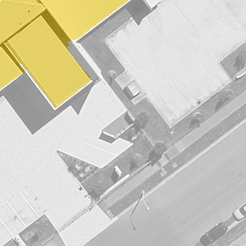,width=\linewidth}}
        
        \vspace{\x}
        
        \centering{\epsfig{figure=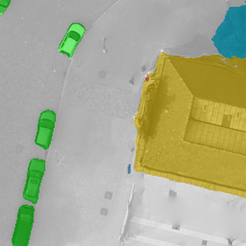,width=\linewidth}}
        
        \vspace{\x}
        
        \centering{\epsfig{figure=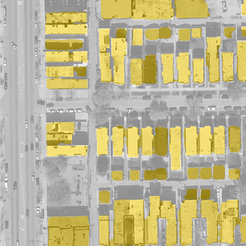,width=\linewidth}}
        
        \vspace{\x}
        
        \centering{\epsfig{figure=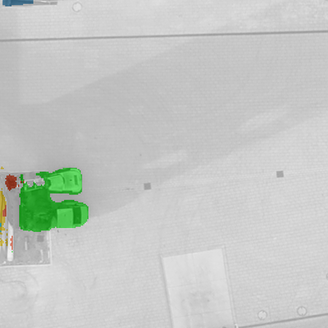,width=\linewidth}}
        
        \captionsetup{width=.95\linewidth}
        DISCA output
    \end{subfigure}\hfill 
    \begin{subfigure}[t]{.19\linewidth}
        \centering{\epsfig{figure=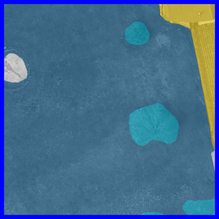,width=\linewidth}}
        
         \vspace{\x}
        
        \centering{\epsfig{figure=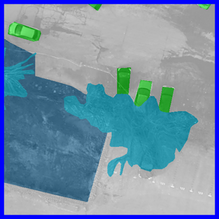,width=\linewidth}}
        
        \vspace{\x}
        
        \centering{\epsfig{figure=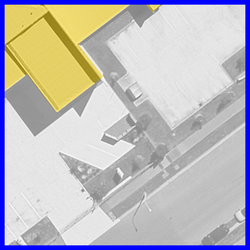,width=\linewidth}}
        
        \vspace{\x}
        
        \centering{\epsfig{figure=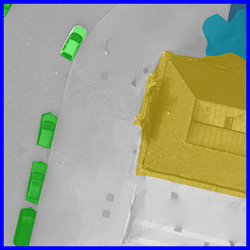,width=\linewidth}}
        
        \vspace{\x}
        
        \centering{\epsfig{figure=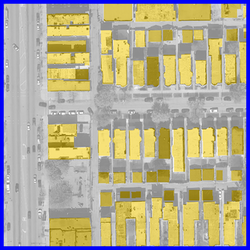,width=\linewidth}}
        
        \vspace{\x}
        
        \centering{\epsfig{figure=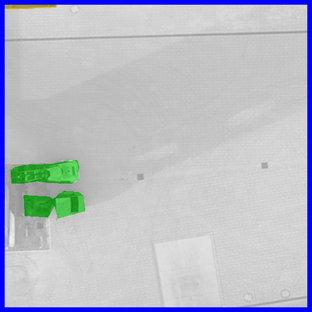,width=\linewidth}}
        
        % \captionsetup{width=.95\linewidth}
        Ground-truth
    \end{subfigure}\hfill
    \caption{Initial output corrected with annotations relying on entropy. On the "\textit{Entropy}" column, the areas with an entropy higher than the ninth quantile over the image are highlighted in red. On the "\textit{Annotation(s)}" column, the color of the annotations represents their labels w.r.t. the associated ground-truth maps.} % Annotations based on uncertainty measurements.
    \label{fig:uncertainty_examples}
\end{figure*}